\documentclass[journal]{IEEEtran}
\usepackage{commath,graphicx,afterpage,subfigure,amsmath,amsfonts,amssymb,algorithm,algorithmic}
\usepackage{graphicx,times,amssymb,cite,subfigure,stfloats,booktabs,multirow,bm, pifont}
\usepackage{cite,multirow,bm,multicol,booktabs,threeparttable,extpfeil}
\usepackage{graphicx}
\usepackage{url}

\begin{document}

\title{DBConformer: Dual-Branch Convolutional Transformer for EEG Decoding}

\author{Ziwei~Wang, Hongbin~Wang, Tianwang~Jia, Xingyi~He, Siyang~Li, and Dongrui~Wu$^{\ast}$, \IEEEmembership{Fellow,~IEEE}
\thanks{This research was supported by Major Science and Technology Project of Industry-university-research Cooperation for Colleges and Universities in Hebei Province in Shijiazhuang City 2511303107A.}
\thanks{Z.~Wang, H.~Wang, T.~Jia, X.~He, S.~Li, and D.~Wu are with the Ministry of Education Key Laboratory of Image Processing and Intelligent Control, School of Artificial Intelligence and Automation, Huazhong University of Science and Technology, Wuhan 430074, China.}
\thanks{*Corresponding Author: Dongrui Wu (drwu09@gmail.com).}}

\markboth{}
{Shell \MakeLowercase{Wang~\emph{et al.}}: }
\maketitle

\begin{abstract}
Electroencephalography (EEG)-based brain-computer interfaces (BCIs) transform spontaneous/evoked neural activity into control commands for external communication. While convolutional neural networks (CNNs) remain the mainstream backbone for EEG decoding, their inherently short receptive field makes it difficult to capture long-range temporal dependencies and global inter-channel relationships. Recent CNN-Transformer (Conformer) hybrids partially address this issue, but most adopt a serial design, resulting in suboptimal integration of local and global features, and often overlook explicit channel-wise modeling. To address these limitations, we propose DBConformer, a dual-branch convolutional Transformer network tailored for EEG decoding. It integrates a temporal Conformer to model long-range temporal dependencies and a spatial Conformer to extract inter-channel interactions, capturing both temporal dynamics and spatial patterns in EEG signals. A lightweight channel attention module further refines spatial representations by assigning data-driven importance to EEG channels. Extensive experiments under four evaluation settings on three paradigms, including motor imagery, seizure detection, and steady-state visual evoked potential, demonstrated that DBConformer consistently outperformed 13 competitive baseline models, with over an eight-fold reduction in parameters than current high-capacity EEG Conformer architecture. Furthermore, the visualization results confirmed that the features extracted by DBConformer are physiologically interpretable and aligned with prior knowledge. The superior performance and interpretability of DBConformer make it reliable for accurate, robust, and explainable EEG decoding. Code is publicized at \url{https://github.com/wzwvv/DBConformer}.
\end{abstract}


\section{Introduction}
A brain-computer interface (BCI) provides a direct communication pathway between cortical activity and external devices \cite{Rosenfeld2017}. Non-invasive BCIs based on electroencephalography (EEG) have gained prominence because EEG sensors are inexpensive, portable, and safe for extended use \cite{wu2020transfer}. Nevertheless, the low signal-to-noise ratio, non-stationarity, and significant inter-subject variability of scalp EEG impose stringent requirements on the decoding model that translates raw signals into task-relevant information.

This study focuses on three significant EEG decoding tasks:
\begin{enumerate}
\item Motor imagery (MI) classification \cite{Pfurtscheller2001}, which enables brain-based control of devices/prostheses. During MI, users imagine the movement of specific body parts, producing event-related desynchronization/synchronization patterns over the brain’s motor cortex \cite{Wu2022NN}.
\item Epileptic seizure detection \cite{Acharya2013}, a task of automated neurological monitoring. Epilepsy, affecting over 70 million people worldwide \cite{thijs2019epilepsy}, is characterized by recurrent seizures. Continuous EEG monitoring remains the gold standard for seizure detection.
\item Steady-state visual evoked potential (SSVEP) \cite{Chen2015} recognition, which occurs when users fixate on flickering visual stimuli, producing EEG responses at the same stimulus frequency and its harmonics.
\end{enumerate}

The decoding model is essential for effective EEG decoding and has been a research hotspot of BCIs. Three trends have emerged in the design of EEG decoding models:
\begin{enumerate}
\item \textit{Model Type:} Classic convolutional neural networks (CNNs), such as EEGNet \cite{Lawhern2018EEGNet}, Shallow CNN (SCNN), and Deep CNN (DCNN) \cite{deepshallow2017}, exploit local spatio-temporal patterns but struggle with global temporal dependencies. CNN-Transformer (Conformer) hybrids extend CNN locality with self-attention based global context. Yet, prior work  \cite{song2022eeg,zhao2024ctnet,ding2024deformer,Li2025TSST}  stacks the two modules serially, restricting interaction between local and global features and neglecting explicit spatial modeling.
\item \textit{Architecture Parallelism and Depth:} Networks evolve from shallow, single-branch (e.g., EEGNet) to deeper, multi-branch designs with more parameters and layers that process frequency bands or multi-resolution inputs in parallel (e.g., FBCNet \cite{mane2021fbcnet}, IFNet \cite{wang2023ifnet}, and ADFCNN \cite{tao2023adfcnn}), enhancing feature diversity.
\item \textit{Multi-View Information Fusion:} Recent studies consider incorporating temporal, spatial, or spectral domain priors explicitly, such as EEG data augmentation \cite{wang2025mvcnet,Wang2024}, time-frequency transforms \cite{dong2025noise,Wang2025CSDA}, frequency band modeling \cite{wang2023ifnet,liu2022fbmsnet}, and spatial filtering initialization \cite{jiang2024csp}. However, existing approaches typically consider these domain priors separately rather than jointly.
\end{enumerate}

Overall, the latest serial Conformer hybrids first compress EEG trials into patch embeddings by CNN blocks and then utilize self-attention to capture global long-range dependencies. Although they are effective for certain tasks, this type of architecture forces all spatial information to pass through a temporal bottleneck, preventing interaction between local and global characteristics. Notably, some Conformer hybrids have large model sizes, e.g., EEG Conformer has around 0.8M parameters, which increases computational cost and may overfit on limited EEG data. Besides, spatial inter-channel relationships, often captured in early CNN layers, may be overwhelmed in later blocks, limiting the model’s capacity to leverage spatial information crucial for MI classification and seizure detection. A balanced, parallel design that treats temporal and spatial patterns symmetrically remains absent.

\begin{figure}[htpb] \centering
\includegraphics[width=\linewidth,clip]{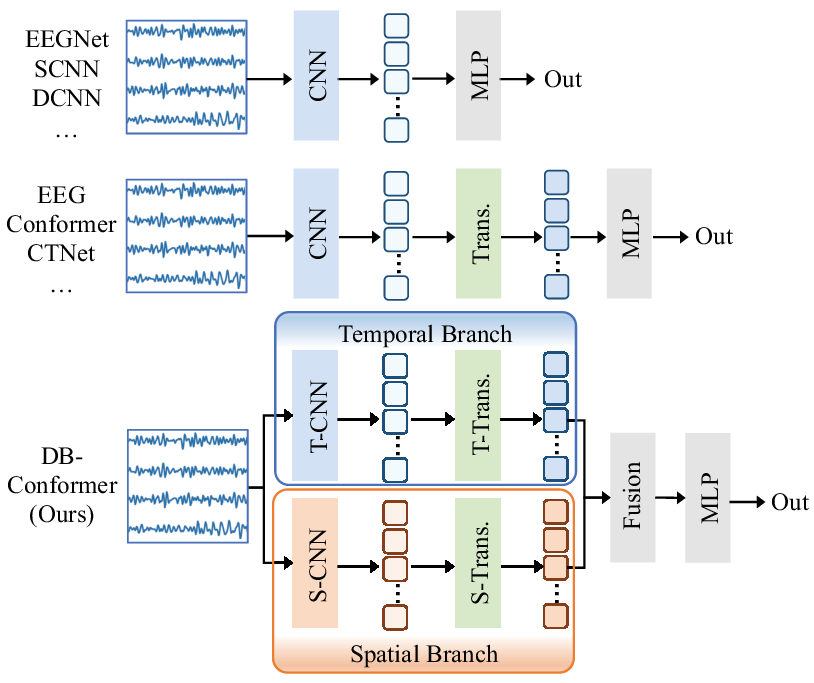}
\caption{Comparison of network architectures among CNNs (EEGNet\cite{Lawhern2018EEGNet}, SCNN \cite{deepshallow2017}, DCNN \cite{deepshallow2017}, etc), traditional serial CNN-Transformer hybrids (EEG Conformer \cite{song2022eeg}, CTNet \cite{zhao2024ctnet}, etc), and the proposed DBConformer. DBConformer has two branches that parallel capture temporal and spatial characteristics.} \label{fig:intro}
\end{figure}

To address the above limitations, we propose DBConformer, a dual-branch convolutional Transformer architecture that performs parallel spatio-temporal representation learning for EEG decoding. The temporal Conformer (T-Conformer) extracts hierarchical temporal dynamics by integrating depth-wise temporal convolutions with multi-head self-attention, whereas the spatial Conformer (S-Conformer) utilizes point-wise convolutions and channel-wise self-attention to model spatial dependencies. A lightweight channel attention module adaptively refines spatial saliency before branch fusion. In each branch, CNN extracts temporal/spatial local representations, and Transformer captures global dependencies in the temporal or channel horizon. The comparison of network architectures among CNNs, traditional serial Conformers, and the proposed DBConformer, illustrated in Fig. \ref{fig:intro}.

The main contributions of this work are summarized as follows:
\begin{enumerate}
\item We propose DBConformer, a dual-branch convolutional Transformer network for EEG decoding. It incorporates a temporal branch T-Conformer and a spatial branch S-Conformer, enabling simultaneous modeling of temporal dynamics and spatial patterns in EEG signals.
\item A lightweight, plug-and-play channel attention module is introduced to learn the relative importance of each EEG channel in a data-driven manner. This module re-weights spatial features before fusion, enhancing both classification accuracy and interpretability without requiring additional supervision.
\item Extensive experiments on three paradigms validated that DBConformer almost always outperforms 13 competitive baseline models. Furthermore, it achieves this superior performance with over an eight-fold reduction in parameters than the high-capacity EEG Conformer baseline, demonstrating both effectiveness and efficiency.
\item Beyond achieving state-of-the-art performance, DBConformer exhibits biologically interpretability, i.e., its learned channel attention scores consistently emphasize sensorimotor-related electrodes (e.g., C3, Cz, C4), aligning well with known neurophysiological priors.
\end{enumerate}

The remainder of this paper is organized as follows: Section~\ref{sect:rw} reviews related works. Section~\ref{sect:me} details the proposed DBConformer. Section~\ref{sect:er} discusses the experimental results and provides analyses. Section~\ref{sect:conclusions} draws conclusions.

\section{Related Works}\label{sect:rw}
EEG decoding has progressed from traditional hand-crafted pipelines to large-scale, end-to-end deep learning architectures. Recent advancements focus on the model design, including CNN-based, CNN-Transformer hybrid, and the emerging CNN-Mamba architectures. A review of current EEG decoding models is shown in Table \ref{tab:model_info}.

Early approaches to MI classification heavily relied on spatial filtering \cite{Blankertz2008} to maximize inter-class variance. In automatic seizure detection, prior features were manually extracted and classified by classifiers such as support vector machines \cite{fu2014classification} and random forest\cite{mursalin2017automated}. These pipelines required substantial domain expertise and manual effort for feature extraction, motivating a shift toward end-to-end deep learning models.

\begin{table*}[htbp]
\centering
\caption{A review of current EEG decoding models.}
\centering\setlength{\tabcolsep}{1mm}
\renewcommand{\arraystretch}{1}
\begin{tabular}{c|ccccc}
\toprule
Model Type & Model Name & Model Structure & \# Branches & Explicit Information Fusion & Paradigms \\
\midrule
\multirow{8}{*}{CNN}
& EEGNet \cite{Lawhern2018EEGNet} & Serial & 1 & No & MI \\
& SCNN \cite{deepshallow2017} & Serial & 1 & No & MI \\
& DCNN \cite{deepshallow2017} & Serial & 1 & No & MI \\
& FBCNet \cite{mane2021fbcnet} & Parallel & 9 & Spectral & MI \\
& FBMSNet \cite{liu2022fbmsnet} & Parallel & 4 & Spectral & MI \\
& IFNet \cite{wang2023ifnet} & Parallel & 2 & Spectral & MI \\
& ADFCNN \cite{tao2023adfcnn} & Parallel & 2 & No & MI \\
& EEGWaveNet \cite{EEGWaveNet2022} & Parallel & 5 & No & Seizure detection \\
\midrule
\multirow{6}{*}{CNN-Mamba}
& \multirow{2}{*}{EEGMamba \cite{gui2024eegmamba}} & \multirow{2}{*}{Serial} & \multirow{2}{*}{1}  & \multirow{2}{*}{No} & MI, seizure detection, sleep stage\\
& & & & & classification, and emotion recognition \\
& MI-Mamba \cite{guo2025mi} & Serial & 1 & No & MI \\
& EEG VMamba \cite{deng2024eegvmamba} & Serial & 1 & No & Seizure prediction \\
& SlimSeiz \cite{lu2024slimseiz} & Serial & 1 & No & Seizure prediction \\
& BiT-MamSleep \cite{zhou2024bit} & Serial & 1 & No & Sleep stage classification \\
\midrule
\multirow{15}{*}{CNN-Transformer}
& EEG Conformer \cite{song2022eeg} & Serial & 1 & No & MI, emotion recognition \\
& CTNet \cite{zhao2024ctnet} & Serial & 1 & No & MI \\
& EEG-Deformer \cite{ding2024deformer} & Serial & 1 & No & Event-related potentials \\
& SE-TSS-Transformer \cite{Li2025TSST} & Serial & 1 & No & Seizure detection \\
& DFformer \cite{kim2024dfformer} & Serial & 1 & No & MI, sleep stage classification \\
& MI-CAT \cite{zhang2023micat} & Serial & 1 & No & MI \\
& MSCFormer \cite{zhao2025multi} & Parallel & 3 & Temporal & MI \\
& TMSA-Net \cite{zhao2025tmsa} & Parallel & 2 & Temporal & MI \\
& MSVTNet \cite{liu2024msvtnet} & Parallel & 4 & Temporal & MI \\
& Dual-TSST \cite{li2025dual} & Parallel & 2 & Temporal & MI, emotion recognition \\
& GAT \cite{song2023global} & Serial & 1 & No & MI \\
& MVCNet \cite{wang2025mvcnet} & Parallel & 2 & Temporal, spatial, and spectral & MI \\
\cmidrule{2-6}
& DBConformer (Ours) & Parallel & 2 & Temporal, spatial & MI, seizure detection \\
\bottomrule
\end{tabular} \label{tab:model_info}
\end{table*}

\subsection{CNNs}
CNNs are well-suited for modeling the local spatio-temporal characteristics of EEG data. Lightweight CNN models such as EEGNet \cite{Lawhern2018EEGNet}, SCNN \cite{deepshallow2017}, FBCNet \cite{mane2021fbcnet}, and IFNet \cite{wang2023ifnet} achieve competitive performance with relatively few parameters. FBMSNet \cite{liu2022fbmsnet} employs multi-scale convolutional blocks to enlarge the receptive field by parallelizing convolutions across different frequency bands. EEGWaveNet \cite{EEGWaveNet2022} introduces a multi-scale temporal CNN architecture with depthwise filters that are channel-specific, extracting multi-scale features from trials of each EEG channel for seizure detection. However, the inherently local nature of convolutional kernels limits their ability to capture long-range temporal dependencies and global spatial patterns.

\subsection{Serial CNN-Transformer Hybrids}
To incorporate global information, recent EEG decoding architectures incorporate Transformers into CNNs, typically following a serial architecture. Specifically, EEG Conformer \cite{song2022eeg}, CTNet \cite{zhao2024ctnet}, EEG-Deformer \cite{ding2024deformer}, SE-TSS-Transformer \cite{Li2025TSST}, and DFformer \cite{kim2024dfformer} first compress EEG trials into patch embeddings using convolutional blocks, and subsequently apply Transformer layers to capture long-range dependencies. In particular, MI-CAT \cite{zhang2023micat} extracts intra-domain and inter-domain features through a temporal-spatial CNN module followed by stacked Transformer blocks. GAT \cite{song2023global} combines parallel convolutions with an attention adaptor to enhance domain transferability. Similarly, MSCFormer \cite{zhao2025multi} and TMSA-Net \cite{zhao2025tmsa} integrate multi-branch, multi-scale CNNs with Transformer encoders to jointly model local and global EEG representations. MSVTNet \cite{liu2024msvtnet} further extracts local spatio-temporal features at different filtered scales and introduces an auxiliary branch loss to facilitate effective CNN-Transformer integration.

Other works employ additional data transformations. For instance, Dual-TSST \cite{li2025dual} applies a wavelet transform to EEG signals and then extracts temporal-spatial-spectral representations from both raw and transformed data via dual CNN branches, followed by Transformer-based fusion. MVCNet \cite{wang2025mvcnet} is a multi-view contrastive network utilizing multi-domain EEG data augmentations and incorporating both cross-view and cross-model modules for MI decoding.

However, this serial design constrains information flow to a single pathway, where local and global features are processed sequentially. As a result, spatial inter-channel relationships, often captured in early CNN layers, may be overwhelmed in later blocks, limiting the model’s capacity to leverage spatial information. Neuroscientific studies further suggest that temporal and spatial information are processed by partly dissociable mechanisms in the brain \cite{marcos2017independent,faugeras2016dissociating}, reinforcing the motivation for architectures that explicitly decouple and then integrate temporal and spatial representations.

\subsection{Other Architectures}
Mamba \cite{gu2023mamba}, a recently proposed alternative to Transformers, provides efficient modeling of long-range dependencies with reduced computational cost. EEGMamba \cite{gui2024eegmamba} and MI-Mamba \cite{guo2025mi} integrate Mamba blocks following CNN layers, adhering to the same serial structure of CNN-Transformer hybrids. However, parallel architectures that jointly model temporal and spatial patterns remain underexplored.

Some research explores architectures beyond CNNs, including full self-attention mechanisms \cite{xie2022transformer,hameed2024temporal}, graph neural networks \cite{ju2023graph,cai2022motor,zhang2020motor,sait2025lightweight}, and long short term memory layers \cite{ahmed2024cardioguard}. Transformer-based models capture long-range dependencies, and graph neural networks leverage electrode topology to refine spatial characteristics. These models are typically parameter-intensive and rarely integrate local feature extraction with global modeling.

Overall, CNNs excel at modeling local spatio-temporal features. CNN-Transformer and CNN-Mamba hybrids offer broader temporal modeling but are limited by the serial design, overlooking the interaction between local and global representations, as well as the spatial patterns.

\begin{figure*}[htpb] \centering
\includegraphics[width=\linewidth,clip]{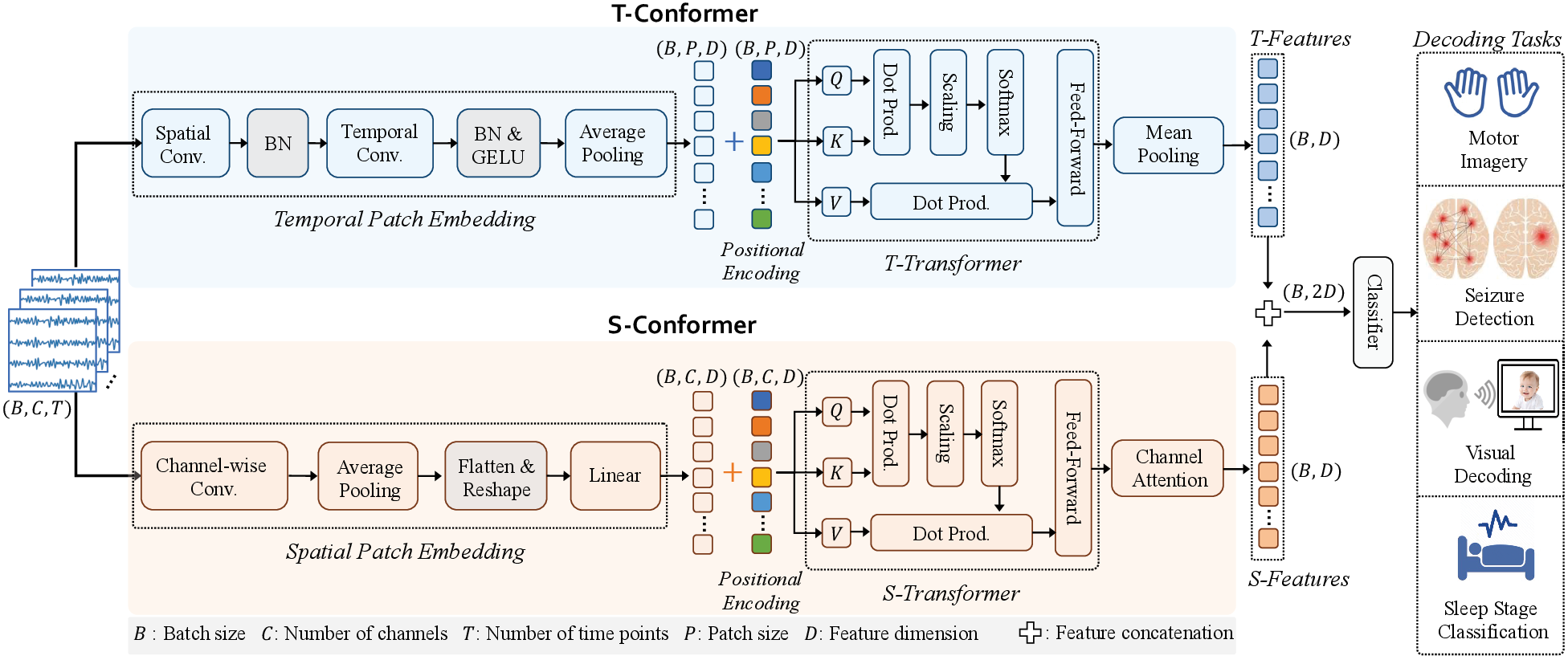}
\caption{Architecture of the proposed DBConformer for EEG decoding.} \label{fig:Framework}
\end{figure*}

\section{DBConformer} \label{sect:me}

\subsection{Overview}
This section details the proposed DBConformer, illustrated in Fig.~\ref{fig:Framework}. In contrast to conventional single-branch EEG decoding networks, such as CNN-based models or serial CNN-Transformer hybrids, DBConformer adopts a parallel dual-branch structure to better leverage the advantages of CNN and Transformer. Specifically, T-Conformer captures long-range temporal dependencies, and S-Conformer models inter-channel relations. A lightweight channel attention module is further incorporated to enhance spatial representations by assigning adaptive, data-driven weights to individual EEG channels. The outputs from both branches are concatenated and fed into the classifier to obtain the final predictions.

\subsection{Data Normalization}
The raw EEG trials of a batch are $\mathbf X\!\in\!\mathbb R^{B\times C\times T}$, where $B$ is the batch size, $C$ the number of channels, and $T$ the number of time samples. We apply Euclidean alignment (EA) \cite{He2020EA} to align the original trials from each subject in the Euclidean space. Assume a subject has $n$ trials, EA can be formulated as the following for the $i$-th trial:
\begin{align}
\tilde{X}_{i}=\bar{R}^{-1 / 2} X_{i},
\end{align}
where $\bar{R}$ is the arithmetic mean of all covariance matrices from a subject:
\begin{align}
\bar{R}=\frac{1}{n} \sum_{i=1}^{n} X_{i} X_{i}^{T}.
\end{align}

After EA, data distribution discrepancies from different subjects are effectively decreased.

\subsection{Network Architecture}
The detailed architecture of DBConformer is summarized in Table~\ref{tab:dbconformer_network}, outlining the composition of each module. As shown in Fig. \ref{fig:Framework} and Table \ref{tab:dbconformer_network}, DBConformer is composed of T-Conformer and S-Conformer branches, along with a classification module. A channel attention module is integrated in S-Conformer for channel-aware weighting.

\begin{table*}[htbp]
\centering
\caption{DBConformer architecture.}
\centering\setlength{\tabcolsep}{0.9mm}
\renewcommand{\arraystretch}{1}
\begin{tabular}{c|c|llllll}
\toprule
Branch & Module & Layer & \# Kernels & Kernel Size & \# Parameters & Output Shape & Options \\
\midrule
\multirow{11.5}{*}{T-Conformer}
& & Separable Conv1D & $F$ & $(1,)$ & $C \cdot F$ & $(B, F, T)$ & Spatial filtering \\
& & BatchNorm1D & -- & -- & $2F$ & $(B, F, T)$ & Normalization \\
& Temporal & Temporal Conv1D & $F$ & $(K,)$ & $F \cdot K$ & $(B, F, T)$ & Temporal depthwise Conv. \\
& Patch & BatchNorm1D \& GELU & -- & -- & $2F$ & $(B, F, T)$ & Normalization and non-linearity \\
& Embedding & Dropout & -- & -- & 0 & $(B, F, T)$ & $p=0.5$ \\
& & AvgPool1D & -- & $(1, W)$ & 0 & $(B, F, P)$ & Patch-wise downsampling \\
& & Permute & -- & -- & 0 & $(B, P, F)$ & Patch tokens along time axis\\
\cmidrule{2-8}
& & Positional Encoding & -- & -- & $P \cdot D$ & $(B, P, D)$ & Learnable temporal encoding \\
& T-Transformer  & Transformer Encoder & -- & -- & $\sim$ & $(B, P, D)$ & $L_t$ layers, $H_t$ heads \\
&  & Mean Pooling & -- & -- & 0 & $(B, D)$ & Mean over patches \\
\midrule
\multirow{12.5}{*}{S-Conformer}
& & Channel-wise Conv1D & 16 & 25 & -- & $(B \cdot C, 16, T')$ & Per-channel temporal filtering \\
& Spatial & AvgPool1D & -- & -- & 0 & $(B \cdot C, 16, 1)$ & Channel-wise downsampling \\
&Patch & Flatten & -- & -- & 0 & $(B \cdot C, 16)$ & Reshape \\
& Embedding& Reshape & -- & -- & 0 & $(B, C, 16)$ & Channel-token preparation \\
& & Linear & 16 & -- & $16D$ & $(B, C, D)$ & Project to embedding space \\
\cmidrule{2-8}
& \multirow{2}{*}{S-Transformer}
& Positional Encoding & -- & -- & $C \cdot D$ & $(B, C, D)$ & Learnable spatial encoding \\
& & Transformer Encoder & -- & -- & $\sim$ & $(B, C, D)$ & $L_s$ layers, $H_s$ heads \\
\cmidrule{2-8}
& & Linear & -- & -- & D $\cdot$ D & $(B, C, D)$ & Intermediate token projection \\
& & Tanh & -- & -- & 0 & $(B, C, D)$ & Non-linearity \\
& Channel & Linear & -- & -- & D & $(B, C, 1)$ & Channel-level scoring\\
& Attention & Softmax & -- & -- & 0 & $(B, C, 1)$ & Normalize across channels\\
& & Weighted Summation & -- & -- & 0 & $(B, D)$ & Weighted channel-aware feature\\
\midrule
\multirow{2}{*}{/}
& Fusion & Concatenation & -- & -- & 0 & $(B, 2D)$ & Feature fusion \\
\cmidrule{2-8}
& Classifier & FC Layers & -- & -- & $2D \rightarrow 64 \rightarrow 32 \rightarrow N_c$ & $(B, N_c)$ & 3-layer MLP (ELU, Dropout) \\
\bottomrule
\end{tabular}

\vspace{0.5em}
\footnotesize{
$B$: batch size,
$C$: number of channels,
$T$: number of time points,
$T'$: temporal length after Conv1D,
$F$: number of filters,
$K$: temporal kernel size,
$W$: pooling window size,
$P = T/W$: number of temporal patches,
$D$: embedding dimension,
$L_t$: number of T-Transformer encoder layers,
$L_s$: number of S-Transformer encoder layers,
$H_t$: number of T-Transformer encoder heads,
$H_s$: number of S-Transformer encoder heads,
$N_c$: number of classes.
}
\label{tab:dbconformer_network}
\end{table*}

\subsubsection{Motivation of Dual-Branch Design}
EEG decoding requires simultaneously modeling fine-grained temporal oscillations and spatially distributed cortical activations. Most existing CNN-Transformer hybrids adopt \emph{serial} designs, where CNN layers first extract embeddings that are subsequently refined by Transformer blocks. Such designs tend to bias the model toward either local feature extraction (when CNN dominates) or global dependency modeling (when Transformer dominates), leading to limited complementarity.

By contrast, DBConformer explicitly decouples temporal and spatial learning into parallel branches with distinct inductive biases:
\begin{itemize}
\item The temporal branch emphasizes local spectral-temporal patterns while also modeling long-range dependencies.
\item The spatial branch emphasizes inter-channel interactions, adaptively weighted by channel attention.
\end{itemize}

Formally, a batch of input EEG trials is denoted as $\mathbf{X} \in \mathbb{R}^{B \times C \times T}$, where $B$ is the number of trials, $C$ the number of channels, and $T$ the number of time samples. DBConformer processes $\mathbf{X}$ through two parallel mappings:
\begin{align}
\mathbf{F}_{\text{t}} &= f_{\text{T-Conformer}}(\mathbf{X}) \in \mathbb{R}^{B \times D}, \\
\mathbf{F}_{\text{s}} &= f_{\text{S-Conformer}}(\mathbf{X}) \in \mathbb{R}^{B \times D}, \\
\mathbf{F}_{\text{fused}} &= [\mathbf{F}_{\text{t}}; \mathbf{F}_{\text{s}}] \in \mathbb{R}^{B \times 2D},
\end{align}
where $D$ is the embedding dimension, $f_{\text{T-Conformer}}(\cdot)$ captures temporal dependencies, and $f_{\text{S-Conformer}}(\cdot)$ captures spatial relations.

This parallel design ensures that temporal and spatial features are learned independently before integration, yielding more disentangled and complementary representations. Unlike prior serial hybrids, this design is further supported by neuroscientific evidence. For example, neurophysiological studies demonstrated that absolute duration and distance are independently coded by distinct populations of prefrontal neurons \cite{marcos2017independent}, whereas behavioral studies showed that temporal and spatial attention recruit separable neural systems \cite{faugeras2016dissociating}. Such findings suggest that temporal rhythms and spatial topographies are processed by distinct but interacting cortical mechanisms, providing both physiological plausibility and theoretical grounding for our dual-branch architecture.

\subsubsection{T-Conformer}
The temporal branch of DBConformer, termed T-Conformer, is designed to capture fine-grained temporal dependencies in EEG trials. It comprises two primary components: a convolutional patch embedding module and a T-Transformer encoder.

\paragraph{Temporal Patch Embedding}
Inspired by prior CNN backbones in EEG decoding \cite{wang2023ifnet}, we design the module with a depthwise separable 1D convolution, a 1D temporal convolution, and an average pooling layer. The first separable convolution has $F$ kernels with a stride of $(1,)$, followed with batch normalization. The second temporal convolution applies $F$ filters, accompanied by batch normalization, a GELU activation, and a dropout with a probability $p=0.5$. Then, an average pooling with kernel size $(1, W)$ is applied along the time dimension to form non-overlapping temporal patches $\mathbf{Z}_{\text{t}}$. Finally, the patches are reshaped via transposition. The number of filters $F$ is set equal to the Transformer embedding dimension $D$, thereby eliminating the need for a separate linear projection.

\paragraph{Temporal Transformer Encoder}
The extracted patches $\mathbf{Z}_{\text{t}}$ pass through a Transformer encoder \cite{vaswani2017attention} to model temporal dependencies. We add a learnable positional encoding $\mathbf{E}_t^{\text{pos}} \in \mathbb{R}^{1 \times P \times D}$ to preserve temporal ordering:
\begin{align}
\mathbf{Z}_t^{\text{in}} = \mathbf{Z}_{\text{t}} + \mathbf{E}_t^{\text{pos}}.
\end{align}

A lightweight Transformer encoder with $L_t$ layers and $H_t$ attention heads is then applied:
\begin{align}
\mathbf{Z}_t^{\text{out}} = \mathrm{TransformerEncoder}(\mathbf{Z}_t^{\text{in}}) \in \mathbb{R}^{B \times P \times D}.
\end{align}

To aggregate temporal patch features into a global representation, we employ mean pooling over the patch dimension:
\begin{align}
\mathbf{F}_{\text{t}} = \frac{1}{P} \sum_{i=1}^{P} \mathbf{Z}_t^{\text{out}}[:, i, :] \in \mathbb{R}^{B \times D},
\end{align}
which is the final temporal representations of input EEG trials.

\subsubsection{S-Conformer}
The spatial branch of DBConformer, termed S-Conformer, is designed to extract inter-channel spatial patterns. It consists of three components: a convolutional spatial patch embedding, an S-Transformer encoder, and a channel attention module.

\paragraph{Spatial Patch Embedding}
This module transforms each EEG trial into a spatial token embedding. A depthwise 1D convolution across the time dimension is designed to extract short-range temporal features per channel. Next, average pooling is applied over the temporal dimension to summarize each channel’s temporal response, yielding a compressed representation. Then, patches are flattened and reshaped to restore batch and channel dimensions. Finally, a linear projection layer is applied to map each channel token to the desired embedding dimension $D$.

\paragraph{Spatial Transformer Encoder}
To capture global spatial dependencies across EEG channels, the embedded channel tokens $\mathbf{Z}_{\text{s}}$ are processed by a lightweight Transformer encoder. A learnable positional encoding $\mathbf{E}_s^{\text{pos}} \in \mathbb{R}^{1 \times C \times D}$ is added to retain channel ordering:
\begin{align}
\mathbf{Z}_s^{\text{in}} = \mathbf{Z}_s + \mathbf{E}_s^{\text{pos}}.
\end{align}

Then, a Transformer encoder with $L_s$ layers and $H_s$ heads is applied:
\begin{align}
\mathbf{Z}_s^{\text{out}} = \mathrm{TransformerEncoder}(\mathbf{Z}_s^{\text{in}}) \in \mathbb{R}^{B \times C \times D}.
\end{align}

\begin{figure}[htpb] \centering
\includegraphics[width=.48\linewidth,clip]{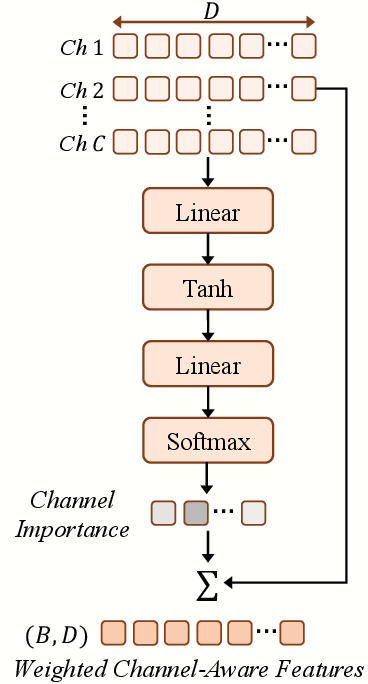}
\caption{The proposed channel attention module.} \label{fig:channel_weight}
\end{figure}

\paragraph{Channel Attention}
A lightweight channel attention module are proposed to adaptively refining the spatial saliency of features from S-Conformer, illustrated in Fig. \ref{fig:channel_weight}.

Specifically, each token $\{z_i^c\}_{c=1}^{C}$ of $\mathbf{Z}_s^{\text{out}}$ is projected through two fully connected layers:
\begin{align}
k_i^c = \mathbf{w}_2^{\top} \tanh(\mathbf{W}_1 z_i^c),
\end{align}
where $\mathbf{W}_1 \in \mathbb{R}^{D \times D}$ and $\mathbf{w}_2 \in \mathbb{R}^{D}$ are learnable parameters, and $tanh(\cdot)$ is the non-linear activation function. The attention weights are then obtained via Softmax normalization:
\begin{align}
\alpha_i^c = \frac{\exp(k_i^c)}{\sum_{j=1}^{C} \exp(k_i^j)}, \quad c = 1, \dots, C.
\end{align}

Then, for each sample, the $c$-th channel token $\mathbf{z}_i^c$ is aggregated through an attention-weighted sum:
\begin{align}
\mathbf{f}_{\text{s}, i} = \sum_{c=1}^{C} \alpha_i^c \cdot \mathbf{z}_i^c.
\end{align}
This operation is executed independently for each sample, yielding its spatial representation $\mathbf{f}_{\text{s}, i} \in \mathbb{R}^{D}$. By repeating this computation across all $B$ samples in a batch, the batch-level spatial representation $\mathbf{F}_{\text{s}} \in \mathbb{R}^{B \times D}$ can be obtained.

\subsubsection{Feature Fusion and Classification}
After obtaining the representations from the temporal and spatial branches, i.e., $\mathbf{F}_{\text{t}} \in \mathbb{R}^{B \times D}$ from T-Conformer and $\mathbf{F}_{\text{s}} \in \mathbb{R}^{B \times D}$ from S-Conformer, the two features are concatenated along the feature dimension to form the fused representation:
\begin{align}
\mathbf{F}_{\text{fused}} = [\mathbf{F}_{\text{t}}; \mathbf{F}_{\text{s}}] \in \mathbb{R}^{B \times 2D},
\end{align}
which is then passed through a multi-layer perceptron classifier to generate the final predictions $\hat{y} = \mathrm{softmax}\big(\mathrm{MLP}(\mathbf{F}_{\text{fused}})\big)$. The classifier consists of three fully connected layers with intermediate ELU activations and dropout regularization.

Given the prediction $\hat{y}_i$ and the ground-truth label $y_i$, the classification loss is:
\begin{align}
\mathcal{L}_{\mathrm{CLS}} = \frac{1}{B} \sum_{i=1}^{B} \mathrm{CE}(\hat{y}_i, y_i), \label{eq:cls_loss}
\end{align}
where $\mathrm{CE}(\cdot,\cdot)$ denotes the cross-entropy loss function and $B$ is the batch size.

\section{Experiments and Results}\label{sect:er}
This section details the datasets, experiments, and analyses. We implemented and fairly evaluated thirteen state-of-the-art EEG decoding models, including CNN-based models, CNN-Transformer hybrids, and CNN-Mamba hybrids. Code for DBConformer, along with all compared baseline models, is available on GitHub\footnote{https://github.com/wzwvv/DBConformer}, serving as a benchmark codebase for EEG decoding.

\subsection{Datasets}
We conducted experiments on two EEG decoding tasks: MI classification and epileptic seizure detection. Characteristics of all datasets are summarized in Table \ref{tab:dataset_info}.

\subsubsection{MI Datasets}
We adopted six publicly available MI datasets from the MOABB benchmark \cite{Sylvain2024} and BCI competitions:
\begin{itemize}
\item BNCI2014001 \cite{tangermann2014001}: Contains EEG trials from 9 subjects performing left/right hand MI, recorded with 22 EEG channels at 250 Hz. The first session was used in the experiments.
\item BNCI2014004 \cite{leeb2014004}: Contains EEG trials from 9 subjects performing left/right hand MI, recorded with only 3 EEG channels at 250 Hz. The first session was used in the experiments.
\item Zhou2016 \cite{Zhou2016}: Contains EEG trials from 4 subjects performing left/right hand MI, recorded with 14 channels at 250 Hz. The first session was used in the experiments.
\item Blankertz2007 \cite{Blankertz2007MI1}: Contains EEG trials from 7 subjects with 59 channels EEG sampled at 250 Hz from BCI Competition IV dataset 1. Subjects 1 and 5 performed left hand/right foot MI, and the other five subjects performed left/right hand MI.
\item BNCI2014002 \cite{steyrl2014002}: Contains EEG trials from 14 subjects, each performing eight runs right hand/feet MI, using 15 EEG channels recorded at 512 Hz. The first five training runs were used in the experiments.
\item OpenBMI \cite{lee2019openbmi}: Contains EEG trials from 54 subjects, each performing two sessions of left/right hand MI, using 62 channels recorded at 1,000 Hz. The total number of trials is 10,800.
\end{itemize}

\subsubsection{Seizure Detection Datasets}
Two clinical seizure detection datasets are evaluated:
\begin{itemize}
\item CHSZ \cite{Wang2023TASA}: Contains EEG recordings from 27 pediatric patients aged 3 months to 10 years, recorded with 19 unipolar electrodes at a sampling rate of 500 or 1000 Hz. All recordings were manually annotated by neurologists to mark the onset and termination of seizures.
\item NICU \cite{Stevenson2019}: A large-scale neonatal EEG dataset collected from 79 full-term neonates in the Helsinki University Hospital neonatal intensive care unit. Data were recorded at 256 Hz, with the same electrodes as CHSZ. Seizures were independently annotated by three clinical experts with a 1s window. A consensus set comprising 39 neonates with confirmed seizures was utilized in the experiments.
\end{itemize}

\subsubsection{SSVEP Dataset}
The Nakanishi2015 \cite{Nakanishi2015} SSVEP dataset is evaluated. Data were recorded at 256 Hz from 10 subjects, with eight channels PO7, PO3, POZ, PO4, PO8, O1, Oz and O2 in the occipital area. For each subject, 12-class stimuli were recorded with 15 blocks.

\begin{table*}[htpb]  \centering \setlength{\tabcolsep}{1.5mm}
\renewcommand{\arraystretch}{1}
\caption{Summary of the six MI datasets, two seizure datasets, and one SSVEP dataset.}
\footnotesize
\label{tab:dataset_info}
\begin{tabular}{c|c|c|c|c|c|c|c}
\toprule
\multirow{2}{*}{Paradigm} & \multirow{2}{*}{Dataset} & Number of & Number of & Sampling & Trial Length & Number of & \multirow{2}{*}{Task Types} \\
& & Subjects & EEG Channels & Rate (Hz) & (seconds) & Total Trials & \\
\midrule
\multirow{5}{*}{MI}
& BNCI2014001 & 9 & 22 & 250 & 4 & 1,296 & left/right hand \\
& BNCI2014004 & 9 & 3 & 250 & 5 & 1,080 & left/right hand \\
& Zhou2016 & 4 & 14 & 250 & 5 & 409 & left/right hand\\
& Blankertz2007 & 7 & 59 & 250 & 3 & 1,400 & left/right hand or left hand/right foot \\
& BNCI2014002 & 14 & 15 & 512 & 5 & 1,400 & right hand/both feet \\
& OpenBMI & 54 & 62 & 1000 & 4 & 10,800 & left/right hand \\
\midrule
\multirow{2}{*}{Seizure detection}
& CHSZ & 27 & 19 & 500 or 1,000 & 4 & 21,237 & normal/seizure \\
& NICU & 39 & 19 & 256 & 4 & 52,534 & normal/seizure \\
\midrule
SSVEP & Nakanishi2015 & 10 & 8 & 256 & 1 & 1,800 & 12 different stimuli \\
\bottomrule
\end{tabular}
\end{table*}

\subsubsection{Data Preprocessing}
For the MI paradigm, the standard preprocessing steps in MOABB, including notch filtering, band-pass filtering, etc., were performed to ensure reproducibility on BNCI2014001, BNCI2014004, Zhou2016, and BNCI2014002 datasets. For the Blankertz2007 dataset, EEG trials were first band-pass filtered between 8 and 30 Hz. Then, trials between [0.5, 3.5] seconds after the cue onset were segmented and downsampled to 250 Hz. For the OpenBMI dataset, the raw EEG signals were first filtered by a third-order Butterworth band-pass filter with a passband of [4, 40] Hz to remove low-frequency drifts and high-frequency noise, and then downsampled to 256 Hz. Following previous studies such as FBCNet \cite{mane2021fbcnet} and FBMSNet \cite{liu2022fbmsnet}, 20 motor-related channels \{FC5, FC3, FC1, FC2, FC4, FC6, C5, C3, C1, Cz, C2, C4, C6, CP5, CP3, CP1, CPz, CP2, CP4, CP6\} were selected for analysis, which cover the sensorimotor cortex regions typically associated with MI tasks. EA was utilized after pre-processing on all MI datasets, and the EA reference matrix was updated online when new test trials arrived on-the-fly, as in \cite{Li2024T-TIME}.

For seizure datasets, 18 bipolar channels were first generated from 19 unipolar channels, following \cite{Stevenson2019}. EEG recordings were segmented into 4s non-overlapping trials. Then, trials from both datasets were preprocessed by a 50 Hz notch filter and a 0.5-50 Hz bandpass filter. For CHSZ, 1000 Hz trials were downsampled to 500 Hz for consistency. For NICU, the ground-truth label of each trial was determined by the majority vote of three experts.

For the SSVEP paradigm, EEG trials were first down-sampled to 256 Hz and [6, 80] Hz band-pass filtered, and then split into 1-second length trials in Nakanishi2015, following the preprocessing steps in \cite{Pan2022}.

\subsection{Compared Approaches}

Thirteen EEG decoding models were compared with the proposed DBConformer in the experiments:
\begin{itemize}
\item EEGNet \cite{Lawhern2018EEGNet} is a compact CNN tailored for EEG classification. It includes two key convolutional blocks: a temporal convolution for capturing frequency-specific features, followed by a depthwise spatial convolution. A separable convolution and subsequent pointwise convolution are designed to enhance spatio-temporal representations.
\item SCNN \cite{deepshallow2017} is a shallow CNN inspired by the filter bank common spatial pattern, utilizing temporal and spatial convolutions in two stages to efficiently extract discriminative EEG features.
\item DCNN \cite{deepshallow2017} is a deeper version of SCNN with larger parameters, consisting of four convolutional blocks with max pooling layers.
\item FBCNet \cite{mane2021fbcnet} incorporates spatial convolution and a temporal variance layer to extract spectral-spatial features across multiple EEG frequency bands.
\item ADFCNN \cite{tao2023adfcnn} is a dual-branch CNN model employing multi-scale temporal convolutions for frequency domain analysis and separable spatial convolutions for global spatial feature extraction. Features from both branches are fused using attention-based fusion before being fed into a fully connected classification head.
\item CTNet \cite{zhao2024ctnet} combines a convolutional front-end, similar to EEGNet, with a Transformer encoder for capturing long-range dependencies in EEG data. This serial structure enhances both local and global feature learning.
\item EEG Conformer \cite{song2022eeg} integrates a convolutional block, a Transformer encoder block, and a classifier. The convolutional block is equipped with temporal and spatial convolutional layers, followed by an average pooling layer. CNN, Transformer, and the classifier are serially equipped, similar to CTNet.
\item MSCFormer \cite{zhao2025multi} integrates multi-branch, multi-scale CNN modules with Transformer encoders to jointly model local and global EEG representations.
\item TMSA-Net \cite{zhao2025tmsa} also integrates multi-scale CNNs with the local and global attention modules to enhance the feature extraction.
\item MSVTNet \cite{liu2024msvtnet} extracts local spatio-temporal features at different filtered scales and introduces an auxiliary branch loss to facilitate effective CNN-Transformer integration.
\item SlimSeiz \cite{lu2024slimseiz} integrates multiple 1D CNN layers to extract features at various temporal resolutions and a Mamba block to capture long-range temporal dependencies for seizure prediction. SlimSeiz enhances temporal modeling by integrating the Mamba block while avoiding a significant parameter overhead.
\item IFNet \cite{wang2023ifnet} extends the spectral-spatial approach by decomposing EEG into multiple frequency bands (e.g., 4-16 Hz, 16-40 Hz). For each band, it applies 1D spatial and temporal convolutions, followed by feature concatenation and a fully connected layer for classification.
\item EEGWaveNet \cite{EEGWaveNet2022} introduces a multi-scale temporal CNN architecture with depthwise filters that are channel-specific, extracting multi-scale features from trials of each EEG channel.
\end{itemize}

\subsection{Implementation}

\subsubsection{Evaluation Scenarios}
To evaluate the performance under different conditions, we performed four experimental settings: chronological order (CO), cross-validation (CV), leave-one-subject-out (LOSO), and cross-dataset (CD), covering within-subject, cross-subject, and cross-device settings, as in \cite{wang2025mvcnet,wang2025cst}.
\begin{itemize}
\item CO: Trials from each subject were split according to recording time. The first 80\% trials were used for training, and the remaining 20\% for testing. This scenario reflects a within-subject evaluation protocol that mimics real-time deployment.
\item CV: A 5-fold cross-validation was adopted, where the dataset was divided chronologically into five equal parts with balanced classes. In each iteration, four folds were used for training and one for testing. This also constitutes a within-subject setting, following the protocol in \cite{mane2021fbcnet}.
\item LOSO: EEG trials from a single subject were held out as the test set, while data from all remaining subjects were joint for training. This setting evaluates the model’s cross-subject decoding ability.
\item CD: Cross-dataset validation, where training and testing were performed on distinct EEG datasets, e.g., training on BNCI2014001 and testing on BNCI2014004. The CD results are shown in Table S1 of the supplementary material.
\end{itemize}

\subsubsection{Evaluation Metrics}
Accuracy is employed to evaluate the performance on MI and SSVEP. To evaluate the classification performance on class-imbalanced seizure datasets, three additional metrics, area under the receiver operating characteristic curve (AUC), weighted F$1$ score and balanced classification accuracy (BCA), are included in seizure detection.

\subsubsection{Parameter Settings}
All experiments were repeated five times with a random seed list $[1, 2, 3, 4, 5]$, and the average results were reported.  For all MI datasets, models were trained for 100 epochs using the Adam optimizer with a learning rate $10^{-3}$ and a batch size of 32.  For seizure datasets, models are trained for 200 epochs using the Adam optimizer with a learning rate $10^{-3}$. The batch size was set to 256 for CHSZ and 512 for NICU because of the larger data size. For all baseline models, the model hyperparameters (e.g., kernel sizes, number of convolutional and Transformer layers) were set according to the original papers to ensure a fair comparison and reproducibility. Due to the specific characteristics of datasets, scenarios, and tasks, we adopted distinct model configurations accordingly, as summarized in Table S6 of the supplementary Material.

\subsection{Results on MI Classification}
Table~\ref{tab:mi_results} presents the average classification accuracies of DBConformer and nine baseline models across six MI datasets under three evaluation settings. Observe that:
\begin{itemize}
\item Among the nine baseline models, IFNet achieved the highest average accuracy across all three settings, attributed to its frequency-aware architectural design. EEG Conformer ranked second under the CO and CV settings but exhibits reduced performance in the LOSO scenario, indicating that its large parameter size may cause overfitting when generalizing to unseen subjects.
\item Compared to the CO setting, results under CV were generally higher across all models, consistent with findings in \cite{wang2025mvcnet}, where temporally ordered training limits generalization due to potential signal drift or session variance. CV scenario benefited from more diverse training trials within each subject. This highlights the importance of fair and appropriate partitioning when benchmarking EEG decoding models.
\item In the LOSO setting, DBConformer maintained the best generalization across subjects. This highlights the robustness of DBConformer in cross-subject EEG decoding tasks.
\item DBConformer almost always outperformed all 12 baseline models across all datasets and settings, demonstrating the effectiveness of the dual-branch design in capturing both temporal and spatial characteristics.
\end{itemize}

\begin{table*}[htpb]
\centering
\setlength{\tabcolsep}{0.2mm}
\renewcommand\arraystretch{1.2}
\footnotesize
\caption{Average classification accuracies (\%) of DBConformer and twelve baseline models on six MI datasets under CO, CV, and LOSO scenarios. The best average performance of each dataset is marked in bold, and the second best by an underline.}
\label{tab:mi_results}
\begin{tabular}{c|c|ccccccccccccc}
\toprule
Task & Dataset & EEGNet & SCNN & DCNN & FBCNet & ADFCNN & CTNet & EEG Confor. & MSCFor. & TMSA-Net & MSVTNet & SlimSeiz & IFNet & DBConfor.\\
\midrule
\multirow{7.5}{*}{CO}
& 2014001 & 69.05$_{\pm1.0}$ & 73.57$_{\pm2.4}$ & 59.29$_{\pm1.6}$ & 68.97$_{\pm1.3}$ & 73.73$_{\pm2.3}$ & 73.49$_{\pm2.1}$ & \underline{78.57}$_{\pm0.7}$ & 75.00$_{\pm1.2}$ & 76.67$_{\pm2.1}$ & 74.60$_{\pm3.0}$ & 68.89$_{\pm2.1}$ & 77.94$_{\pm0.9}$ & \textbf{80.16}$_{\pm1.8}$ \\
& 2014004 & 68.43$_{\pm2.2}$ & 68.15$_{\pm1.4}$ & 62.41$_{\pm1.6}$ & 65.46$_{\pm1.6}$ & 69.63$_{\pm1.2}$ & 71.57$_{\pm1.3}$ & 72.04$_{\pm1.6}$ & 72.20$_{\pm1.8}$ & 70.28$_{\pm2.0}$ & 71.30$_{\pm1.5}$ & 68.24$_{\pm2.7}$ & \underline{73.43}$_{\pm1.7}$ & \textbf{75.09}$_{\pm1.5}$ \\
& Zhou2016 & 80.13$_{\pm3.4}$ & 75.03$_{\pm6.2}$ & 78.03$_{\pm2.4}$ & 63.33$_{\pm2.3}$ & 71.42$_{\pm2.0}$ & 76.81$_{\pm5.1}$ & 73.87$_{\pm4.5}$ & 73.93$_{\pm4.5}$ & 67.94$_{\pm3.0}$ & 70.32$_{\pm1.4}$ & 72.01$_{\pm4.0}$ & \underline{81.70}$_{\pm2.1}$ & \textbf{82.49}$_{\pm2.4}$ \\
& Blankertz & 78.79$_{\pm2.8}$ & 76.71$_{\pm2.1}$ & 70.00$_{\pm4.1}$ & 75.93$_{\pm1.3}$ & 76.07$_{\pm1.4}$ & 79.00$_{\pm3.4}$ & 82.29$_{\pm2.6}$ & \underline{85.29}$_{\pm1.6}$ & 78.93$_{\pm1.0}$ & 84.29$_{\pm0.9}$ & 65.64$_{\pm1.2}$& 84.00$_{\pm0.6}$ & \textbf{87.21}$_{\pm1.3}$ \\
& 2014002 & 66.07$_{\pm2.8}$ & \underline{79.07}$_{\pm2.0}$ & 64.07$_{\pm2.7}$ & 69.50$_{\pm1.0}$ & 73.00$_{\pm2.0}$ & 71.00$_{\pm1.8}$ & 76.21$_{\pm1.5}$ & 75.79$_{\pm1.9}$ & 76.79$_{\pm1.9}$ & 78.93$_{\pm1.4}$ & 78.71$_{\pm0.5}$& 78.29$_{\pm1.7}$ & \textbf{79.14}$_{\pm0.2}$ \\
& OpenBMI & 65.54$_{\pm0.7}$ & 71.78$_{\pm0.3}$ & 66.84$_{\pm1.1}$ & 61.31$_{\pm0.6}$ & 71.82$_{\pm0.8}$ & 68.58$_{\pm0.7}$ & \underline{71.84}$_{\pm0.8}$ & 70.64$_{\pm0.6}$ & 71.10$_{\pm0.6}$ & 70.06$_{\pm0.7}$ & 61.15$_{\pm0.4}$ & 68.19$_{\pm0.6}$ & \textbf{72.12}$_{\pm0.1}$ \\
\cmidrule{2-15}
& Average & 71.33 & 74.05 & 66.77 & 67.42 & 72.61 & 73.41 & 75.80 & 75.47 & 73.62 & 74.91 & 69.11 & \underline{77.26} & \textbf{79.37} \\
\midrule
\multirow{7.5}{*}{CV}
& 2014001 & 71.86$_{\pm1.0}$ & 78.22$_{\pm0.2}$ & 61.59$_{\pm1.1}$ & 74.74$_{\pm0.7}$ & 77.27$_{\pm0.6}$ & 75.97$_{\pm0.8}$ & \underline{83.62}$_{\pm1.0}$ & 81.83$_{\pm0.7}$ & 82.32$_{\pm0.5}$ & 81.28$_{\pm0.7}$ & 75.56$_{\pm1.5}$ & 82.62$_{\pm0.9}$ & \textbf{83.66}$_{\pm0.4}$ \\
& 2014004 & 68.91$_{\pm0.9}$ & 68.13$_{\pm0.6}$ & 63.51$_{\pm0.4}$ & 67.17$_{\pm0.3}$ & 69.48$_{\pm1.2}$ & 70.96$_{\pm1.1}$ & 73.15$_{\pm0.7}$ & 72.30$_{\pm0.4}$ & 72.11$_{\pm0.9}$ & \underline{73.59}$_{\pm0.6}$ & 69.74$_{\pm0.5}$ & 72.02$_{\pm0.5}$ & \textbf{74.17}$_{\pm0.6}$ \\
& Zhou2016 & 85.30$_{\pm2.1}$ & 82.90$_{\pm0.9}$ & 79.68$_{\pm1.1}$ & 80.01$_{\pm0.5}$ & 84.95$_{\pm1.5}$ & 86.38$_{\pm1.5}$ & 85.67$_{\pm1.5}$ & \underline{89.16}$_{\pm0.8}$ & 85.54$_{\pm1.3}$ & 88.06$_{\pm1.2}$ & 85.60$_{\pm1.3}$ & 87.78$_{\pm1.0}$ & \textbf{91.54}$_{\pm0.2}$ \\
& Blankertz & 80.21$_{\pm0.8}$ & 81.11$_{\pm0.7}$ & 69.83$_{\pm1.5}$ & 84.14$_{\pm0.6}$ & 81.10$_{\pm1.8}$ & 83.64$_{\pm1.0}$ & 87.43$_{\pm0.3}$ & 88.41$_{\pm0.9}$ & 74.09$_{\pm0.6}$ & \underline{89.07}$_{\pm0.8}$ & 75.37$_{\pm1.1}$ & 88.23$_{\pm0.5}$ & \textbf{90.33}$_{\pm0.5}$ \\
& 2014002 & 68.90$_{\pm1.4}$ & \underline{81.24}$_{\pm0.5}$ & 67.17$_{\pm0.3}$ & 74.53$_{\pm0.7}$ & 74.63$_{\pm0.8}$ & 75.50$_{\pm0.9}$ & 79.30$_{\pm0.3}$ & 80.29$_{\pm0.7}$ & 80.41$_{\pm0.6}$ & 80.91$_{\pm0.5}$ & 77.93$_{\pm0.9}$ & 80.21$_{\pm0.5}$ & \textbf{81.36}$_{\pm0.3}$ \\
& OpenBMI & 65.99$_{\pm0.4}$ & 71.99$_{\pm0.3}$ & 67.68$_{\pm0.4}$ & 63.82$_{\pm0.2}$ & 72.21$_{\pm0.5}$ & 69.50$_{\pm0.4}$ & \underline{72.34}$_{\pm0.3}$ & 71.34$_{\pm0.2}$ & 72.07$_{\pm0.2}$ & 71.20$_{\pm0.2}$ & 61.23$_{\pm0.5}$ & 71.75$_{\pm0.2}$ & \textbf{72.89}$_{\pm0.2}$ \\
\cmidrule{2-15}
& Average & 73.53 & 77.27 & 68.24 & 74.07 & 76.61 & 76.99 & 80.25 & 80.55 & 77.76 & \underline{80.69} & 74.24 & 80.44 & \textbf{82.32} \\
\midrule
\multirow{7.5}{*}{LOSO}
& 2014001 & 73.64$_{\pm1.1}$ & 72.22$_{\pm1.0}$ & 73.21$_{\pm1.8}$ & 72.56$_{\pm1.0}$ & 71.76$_{\pm0.8}$ & 73.40$_{\pm1.2}$ & 73.07$_{\pm2.0}$ & 76.02$_{\pm0.8}$ & \underline{77.32}$_{\pm0.5}$ & 77.10$_{\pm1.3}$ & 69.62$_{\pm1.3}$ & 74.52$_{\pm0.7}$ & \textbf{77.67}$_{\pm0.6}$ \\
& 2014004 & \underline{67.78}$_{\pm0.9}$ & 62.19$_{\pm0.8}$ & 62.56$_{\pm0.6}$ & 67.09$_{\pm0.7}$ & 65.17$_{\pm0.7}$ & 65.28$_{\pm1.1}$ & 64.22$_{\pm1.1}$ & 64.72$_{\pm0.6}$ & 66.02$_{\pm0.8}$ & 67.41$_{\pm0.8}$ & 63.15$_{\pm1.4}$ & 67.69$_{\pm0.4}$ & \textbf{69.85}$_{\pm0.5}$ \\
& Zhou2016 & 83.22$_{\pm1.8}$ & 82.10$_{\pm0.7}$ & 83.84$_{\pm1.2}$ & 82.07$_{\pm0.9}$ & 82.09$_{\pm1.3}$ & 83.88$_{\pm1.0}$ & 82.43$_{\pm1.5}$ & 85.31$_{\pm2.4}$ & 80.78$_{\pm2.3}$ & 84.95$_{\pm1.4}$ & 84.06$_{\pm1.6}$ & \textbf{86.21}$_{\pm1.0}$ & \underline{85.37}$_{\pm0.9}$ \\
& Blankertz & 71.10$_{\pm0.8}$ & 70.64$_{\pm0.6}$ & 72.19$_{\pm0.9}$ & \underline{76.23}$_{\pm1.4}$ & 70.59$_{\pm1.7}$ & 69.50$_{\pm1.8}$ & 74.41$_{\pm1.1}$ & 73.66$_{\pm0.9}$ & 72.72$_{\pm0.9}$ & 75.40$_{\pm1.3}$ & 73.21$_{\pm0.8}$ & 73.43$_{\pm1.1}$ & \textbf{76.33}$_{\pm0.7}$ \\
& 2014002 & 72.86$_{\pm0.4}$ & 70.57$_{\pm1.4}$ & \underline{74.34}$_{\pm0.8}$ & 71.31$_{\pm0.8}$ & 72.67$_{\pm0.4}$ & 74.14$_{\pm0.8}$ & 72.84$_{\pm1.4}$ & 74.09$_{\pm0.6}$ & 73.99$_{\pm1.1}$ & 74.61$_{\pm1.1}$ & 73.01$_{\pm0.6}$ & 73.90$_{\pm0.7}$ & \textbf{77.17}$_{\pm0.8}$ \\
& OpenBMI & \underline{73.33}$_{\pm0.1}$ & 68.98$_{\pm0.4}$ & 73.27$_{\pm0.2}$ & 60.34$_{\pm0.6}$ & 73.18$_{\pm0.4}$ & 70.05$_{\pm0.1}$ & 70.05$_{\pm0.1}$ & 73.25$_{\pm0.8}$ & 72.79$_{\pm0.7}$ & 72.77$_{\pm0.3}$ & 68.89$_{\pm0.7}$ & 70.03$_{\pm0.5}$ & \textbf{73.61}$_{\pm0.2}$ \\
\cmidrule{2-15}
& Average & 73.65 & 71.12 & 73.23 & 71.60 & 72.58 & 72.71 & 72.84 & 74.51 & 73.94 & \underline{75.37} & 71.99 & 74.30 & \textbf{76.67} \\
\bottomrule
\end{tabular}
\end{table*}

Further, we conducted paired $t$-tests for each comparison and applied Benjamini-Hochberg False Discovery Rate correction to control for multiple comparisons \cite{benjamini1995}. The adjusted $p$-values are summarized in Table~\ref{tab:t_tests}, where values below 0.05 are highlighted in bold to indicate statistical significance. It is evident that the performance improvements of DBConformer over other competing models were almost always statistically significant.

\begin{table*}[htpb]
\centering
\setlength{\tabcolsep}{3mm}
\renewcommand{\arraystretch}{.9}
\caption{Adjust $p$-values between DBConformer and twelve baseline models under CO, CV, and LOSO settings. Statistically significant results ($p < 0.05$) are marked in bold.}
\label{tab:t_tests}
\begin{tabular}{c|c|cccccc}
\toprule
Task & DBConformer vs. & BNCI2014001 & BNCI2014004 & Zhou2016 & Blankerz2007 & BNCI2014002 & OpenBMI \\
\midrule
\multirow{12}{*}{CO}
& EEGNet & $<$ \textbf{0.001} & $<$ \textbf{0.01} & 0.212 & $<$ \textbf{0.05} & $<$ \textbf{0.001} & $<$ \textbf{0.001} \\
& SCNN & $<$ \textbf{0.01} & $<$ \textbf{0.01} & $<$ \textbf{0.05} & $<$ \textbf{0.001} & 0.183 & 0.104 \\
& DCNN & $<$ \textbf{0.001} & $<$ \textbf{0.001} & 0.198 & $<$ \textbf{0.001} & $<$ \textbf{0.001} & $<$ \textbf{0.01} \\
& FBCNet & $<$ \textbf{0.001} & $<$ \textbf{0.01} & $<$ \textbf{0.001} & $<$ \textbf{0.001} & $<$ \textbf{0.001} & $<$ \textbf{0.001} \\
& ADFCNet & $<$ \textbf{0.001} & $<$ \textbf{0.01} & $<$ \textbf{0.01} & $<$ \textbf{0.01} & $<$ \textbf{0.01} & 0.081 \\
& CTNet & $<$ \textbf{0.01} & 0.052 & 0.054 & $<$ \textbf{0.001} & $<$ \textbf{0.01} & $<$ \textbf{0.01} \\
& EEG Conformer & 0.052 & $<$ \textbf{0.01} & $<$ \textbf{0.01} & $<$ \textbf{0.01} & $<$ \textbf{0.05} & $<$ 0.051 \\
& MSCFormer & $<$ \textbf{0.001} & 0.051 & $<$ \textbf{0.05} & 0.13 & $<$ \textbf{0.001} & $<$ \textbf{0.05} \\
& TMSA-Net & $<$ \textbf{0.01} & $<$ \textbf{0.01} & $<$ \textbf{0.01} & $<$ \textbf{0.01} & $<$ \textbf{0.01} & 0.09 \\
& MSVTNet & $<$ \textbf{0.05} & $<$ \textbf{0.01} & $<$ \textbf{0.01} & 0.073 & 0.136 & $<$ \textbf{0.01} \\
& SlimSeiz & $<$ \textbf{0.001} & $<$ \textbf{0.05} & $<$ \textbf{0.05} & $<$ \textbf{0.001} & 0.116 & $<$ \textbf{0.001} \\
& IFNet & $<$ \textbf{0.05} & $<$ \textbf{0.05} & 0.158 & $<$ \textbf{0.05} & 0.128 & $<$ \textbf{0.001} \\
\midrule
\multirow{12}{*}{CV}
& EEGNet & $<$ \textbf{0.001} & $<$ \textbf{0.001} & $<$ \textbf{0.01} & $<$ \textbf{0.001} & $<$ \textbf{0.001} & $<$ \textbf{0.001} \\
& SCNN & $<$ \textbf{0.001} & $<$ \textbf{0.001} & $<$ \textbf{0.001} & $<$ \textbf{0.001} & 0.16 & 0.053 \\
& DCNN & $<$ \textbf{0.001} & $<$ \textbf{0.001} & $<$ \textbf{0.001} & $<$ \textbf{0.001} & $<$ \textbf{0.001} & $<$ \textbf{0.001} \\
& FBCNet & $<$ \textbf{0.001} & $<$ \textbf{0.001} & $<$ \textbf{0.001} & $<$ \textbf{0.001} & $<$ \textbf{0.001} & $<$ \textbf{0.001} \\
& ADFCNet & $<$ \textbf{0.001} & $<$ \textbf{0.01} & $<$ \textbf{0.001} & $<$ \textbf{0.001} & $<$ \textbf{0.001} & 0.132 \\
& CTNet & $<$ \textbf{0.001} & $<$ \textbf{0.01} & $<$ \textbf{0.001} & $<$ \textbf{0.01} & $<$ \textbf{0.001} & $<$ \textbf{0.001} \\
& EEG Conformer & 0.08 & $<$ \textbf{0.01} & $<$ \textbf{0.001} & $<$ \textbf{0.01} & 0.099 & $<$ \textbf{0.05} \\
& MSCFormer & $<$ \textbf{0.01} & $<$ \textbf{0.05} & $<$ \textbf{0.01} & 0.089 & 0.063 & $<$ \textbf{0.05} \\
& TMSA-Net & $<$ \textbf{0.01} & $<$ \textbf{0.01} & $<$ \textbf{0.01} & $<$ \textbf{0.001} & 0.075 & 0.055 \\
& MSVTNet & $<$ \textbf{0.05} & 0.158 & $<$ \textbf{0.01} & 0.094 & 0.051 & $<$ \textbf{0.01} \\
& SlimSeiz & $<$ \textbf{0.001} & $<$ \textbf{0.01} & $<$ \textbf{0.01} & $<$ \textbf{0.001} & $<$ \textbf{0.01} & $<$ \textbf{0.001} \\
& IFNet & $<$ \textbf{0.05} & $<$ \textbf{0.01} & $<$ \textbf{0.001} & $<$ \textbf{0.05} & 0.064 & $<$ \textbf{0.01} \\
\midrule
\multirow{12}{*}{LOSO}
& EEGNet & $<$ \textbf{0.001} & $<$ \textbf{0.05} & 0.082 & $<$ \textbf{0.001} & $<$ \textbf{0.001} & $<$ \textbf{0.05} \\
& SCNN & $<$ \textbf{0.001} & $<$ \textbf{0.001} & $<$ \textbf{0.01} & $<$ \textbf{0.001} & $<$ \textbf{0.001} & $<$ \textbf{0.001} \\
& DCNN & $<$ \textbf{0.001} & $<$ \textbf{0.001} & 0.094 & $<$ \textbf{0.001} & $<$ \textbf{0.01} & $<$ \textbf{0.05} \\
& FBCNet & $<$ \textbf{0.001} & $<$ \textbf{0.001} & $<$ \textbf{0.001} & 0.172 & $<$ \textbf{0.001} & $<$ \textbf{0.001} \\
& ADFCNet & $<$ \textbf{0.001} & $<$ \textbf{0.001} & $<$ \textbf{0.01} & $<$ \textbf{0.001} & $<$ \textbf{0.001} & 0.055 \\
& CTNet & $<$ \textbf{0.001} & $<$ \textbf{0.001} & 0.082 & $<$ \textbf{0.001} & $<$ \textbf{0.01} & $<$ \textbf{0.001} \\
& EEG Conformer & $<$ \textbf{0.001} & $<$ \textbf{0.001} & $<$ \textbf{0.05} & $<$ \textbf{0.05} & $<$ \textbf{0.01} & $<$ \textbf{0.001} \\
& MSCFormer & $<$ \textbf{0.001} & $<$ \textbf{0.001} & 0.172 & $<$ \textbf{0.01} & $<$ \textbf{0.01} & $<$ \textbf{0.05} \\
& TMSA-Net & 0.059 & $<$ \textbf{0.001} & $<$ \textbf{0.05} & $<$ \textbf{0.001} & $<$ \textbf{0.05} & 0.161 \\
& MSVTNet & 0.14 & $<$ \textbf{0.001} & 0.182 & 0.076 & 0.054 & $<$ \textbf{0.05} \\
& SlimSeiz & $<$ \textbf{0.001} & $<$ \textbf{0.001} & $<$ \textbf{0.05} & $<$ \textbf{0.01} & $<$ \textbf{0.001} & $<$ \textbf{0.001} \\
& IFNet & $<$ \textbf{0.001} & $<$ \textbf{0.001} & $<$ \textbf{0.05} & $<$ \textbf{0.001} & $<$ \textbf{0.05} & $<$ \textbf{0.001} \\
\bottomrule
\end{tabular}
\end{table*}

\subsection{Results on Seizure Detection}
For seizure detection, we compared the proposed DBConformer with seven baseline models: EEGNet, SCNN, DCNN, ADFCNN, EEG Conformer, SlimSeiz, and EEGWaveNet. The first five models were originally designed for tasks such as MI and emotion recognition, rather than seizure detection. But they are general for all EEG decoding tasks; thus, we re-implemented and adapted them to seizure detection. SlimSeiz \cite{lu2024slimseiz} is a CNN-Mamba hybrid model for seizure prediction that can be easily implemented in the seizure detection task. EEGWaveNet \cite{EEGWaveNet2022} is a multi-scale depthwise CNN specifically developed for seizure detection, serving as a competitive task-specific benchmark. Table~\ref{tab:seizure_result} summarizes the average AUCs, F1 scores, and BCAs of DBConformer and seven baseline models on the CHSZ and NICU datasets. Observe that:
\begin{itemize}
\item Among the baseline approaches, EEG Conformer and ADFCNN achieved competitive performance across most metrics. EEG Conformer benefits from its temporal modeling capacity, while ADFCNN leverages dual-scale convolutions to capture frequency-sensitive patterns.
\item Compared to CHSZ, all models generally exhibited lower performance on the NICU dataset. This degradation can be attributed to differences in recording conditions, subject age, and seizure morphology. The smaller brain volume of neonates introduces greater variability and reduces the signal-to-noise ratio in NICU, underscoring the need for robust generalization across patients.
\item DBConformer achieved the best performance across all evaluation metrics and both datasets. Compared to strong baselines such as EEG Conformer and ADFCNN, DBConformer yielded a 2-4\% improvement in BCAs. Results demonstrate the adaptability of DBConformer to clinical EEG decoding tasks. The dual-branch spatio-temporal modeling design provides robust representations for varying patients, making it more reliable for real-world seizure detection.
\end{itemize}

\begin{table}[htpb]
\centering
\renewcommand\arraystretch{1}
\caption{Average classification AUCs (\%), F1s (\%), and BCAs (\%) on CHSZ and NICU datasets under the LOSO setting. The best average performance is marked in bold, and the second-best by an underline.}
\label{tab:seizure_result}
\begin{tabular}{c|ccc|ccc}
\toprule
\multirow{2.5}{*}{Model} & \multicolumn{3}{c|}{CHSZ} & \multicolumn{3}{c}{NICU} \\
\cmidrule{2-7}
 & AUC & F1 & BCA & AUC & F1 & BCA \\
\midrule
EEGNet & 84.36 & 87.54  & 74.93 & \underline{71.87} & 71.03 & \underline{64.79} \\
SCNN & 49.17 & 70.18  & 50.15   & 68.86 & 76.54 & 63.52 \\
DCNN & 52.88 & 67.84  & 50.41  & 66.65 & 72.28 & 60.93 \\
ADFCNet & 86.49 & \underline{90.22}  & \underline{76.92} & 70.48 & 69.45 & 62.60 \\
EEG Conformer & \underline{89.23}  & 88.01  & 74.33  & 71.46 & \underline{77.43} & 64.04 \\
SlimSeiz & 86.94 & 87.56 & 76.30 & 71.78 & 70.01 & 64.25 \\
EEGWaveNet & 87.38  & 87.71  & 75.56  & 68.68 & 59.00 & 58.14 \\
DBConformer & \textbf{91.75} & \textbf{90.31} & \textbf{79.01} & \textbf{72.08} & \textbf{79.23} & \textbf{65.21} \\
\bottomrule
\end{tabular}
\end{table}

\subsection{Results on SSVEP Classification}
To further evaluate the generalizability of DBConformer, we conducted experiments on the Nakanishi2015 SSVEP dataset. As shown in Table~\ref{tab:ssvep_results}, DBConformer consistently outperformed recent CNN and CNN-Transformer baselines, including Conformer, MSCFormer, TMSA-Net, and IFNet. In particular, DBConformer achieved the highest average accuracy of 87.59\%, surpassing the second-best IFNet by a substantial margin, highlighting its adaptability across diverse EEG paradigms.

\begin{table}[htpb]
\centering
\setlength{\tabcolsep}{0.9mm}
\renewcommand\arraystretch{1}
\caption{Average classification accuracies (\%) on Nakanishi2015 dataset under the CO setting. The best average performance is marked in bold, and the second-best by an underline.}
\label{tab:ssvep_results}
\begin{tabular}{c|ccccc}
\toprule
Subject & Conformer & MSCFormer & TMSA-Net & IFNet & DBConformer \\
\midrule
S1 & 12.03 & 13.89 & 46.30 & \underline{56.48} & \textbf{62.04}  \\
S2 & 8.33 & 12.96 & 24.07 & \underline{32.41} & \textbf{46.30}  \\
S3 & 8.33 & 13.89 & 57.41 & \underline{70.37} & \textbf{88.89}  \\
S4 & 30.55 & 14.81 & 78.70 & \underline{93.52} & \textbf{96.29}  \\
S5 & 20.37 & 22.22 & 92.59 & \underline{98.15} & \textbf{100.00}  \\
S6 & 35.18 & 15.74 & 87.04 & \underline{92.59} & \textbf{100.00}  \\
S7 & 15.74 & 8.33 & 85.19 & \underline{93.52} & \textbf{100.00}  \\
S8 & 16.66 & 19.44 & 91.67 & \textbf{100.00} & \textbf{100.00}  \\
S9 & 32.41 & 20.37 & 88.89 & \underline{91.67} & \textbf{96.29}  \\
S10 & 10.18 & 11.11 & 54.63 & \underline{65.74} & \textbf{86.11}  \\
\midrule
Average & 18.98 & 15.28 & 70.65 & \underline{79.44} & \textbf{87.59} \\
\bottomrule
\end{tabular}
\end{table}

\subsection{Ablation Study}
\begin{figure}[htpb] \centering
\includegraphics[width=\linewidth,clip]{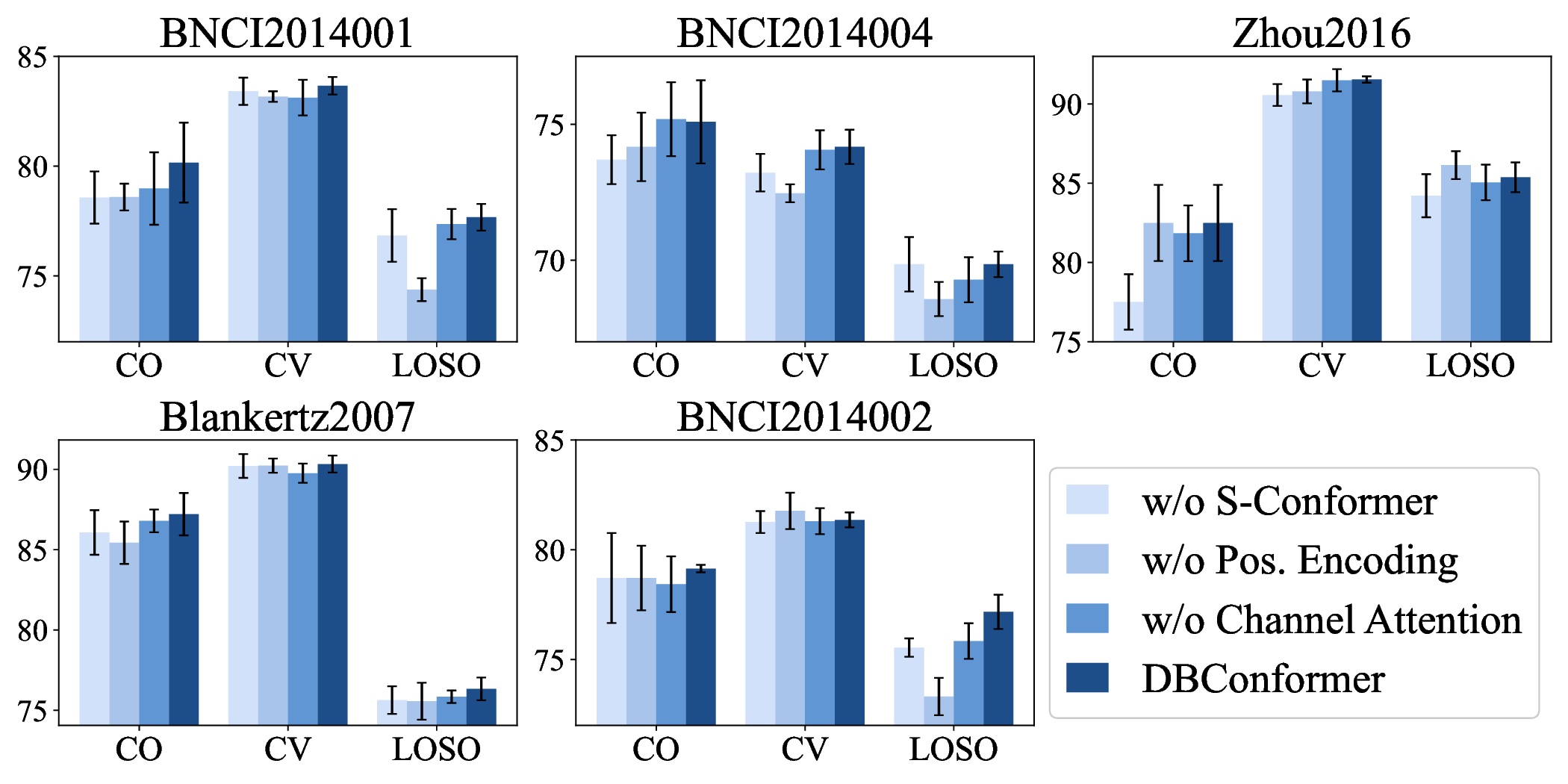}
\caption{Ablation studies on the parallel spatio-temporal design, positional encoding, and the channel attention module.} \label{fig:ablation}
\end{figure}

To further validate the effectiveness of DBConformer's key components, we conducted ablation studies on the parallel spatio-temporal design, positional encoding, and the channel attention module. Experiments were performed on six MI datasets under three settings, shown in Fig. \ref{fig:ablation}. Observe that:
\begin{itemize}
\item \textit{Parallel Spatio-Temporal Modeling.}
We compared the full DBConformer with its T-Conformer-only variant. Adding the S-Conformer branch consistently improved performance across all settings. Results confirmed the effectiveness of incorporating spatial features via the auxiliary S-Conformer, which complemented the dominant temporal modeling of T-Conformer.
\item \textit{Positional Encoding.}
Disabling the learnable positional embeddings for both temporal and spatial tokens resulted in performance drops under all settings. This validated the importance of explicit temporal/channel ordering when modeling patch and channel tokens.
\item \textit{Channel Attention Module.}
Replacing the channel attention module with na\"{\i}ve mean pooling degraded performance, confirming that adaptively reweighting EEG channels benefited enhancing spatial features.
\end{itemize}

\subsection{Effect of Dual-Branch Modeling}
\begin{figure*}[htpb]\centering
\subfigure[]{\includegraphics[width=.48\linewidth,clip]{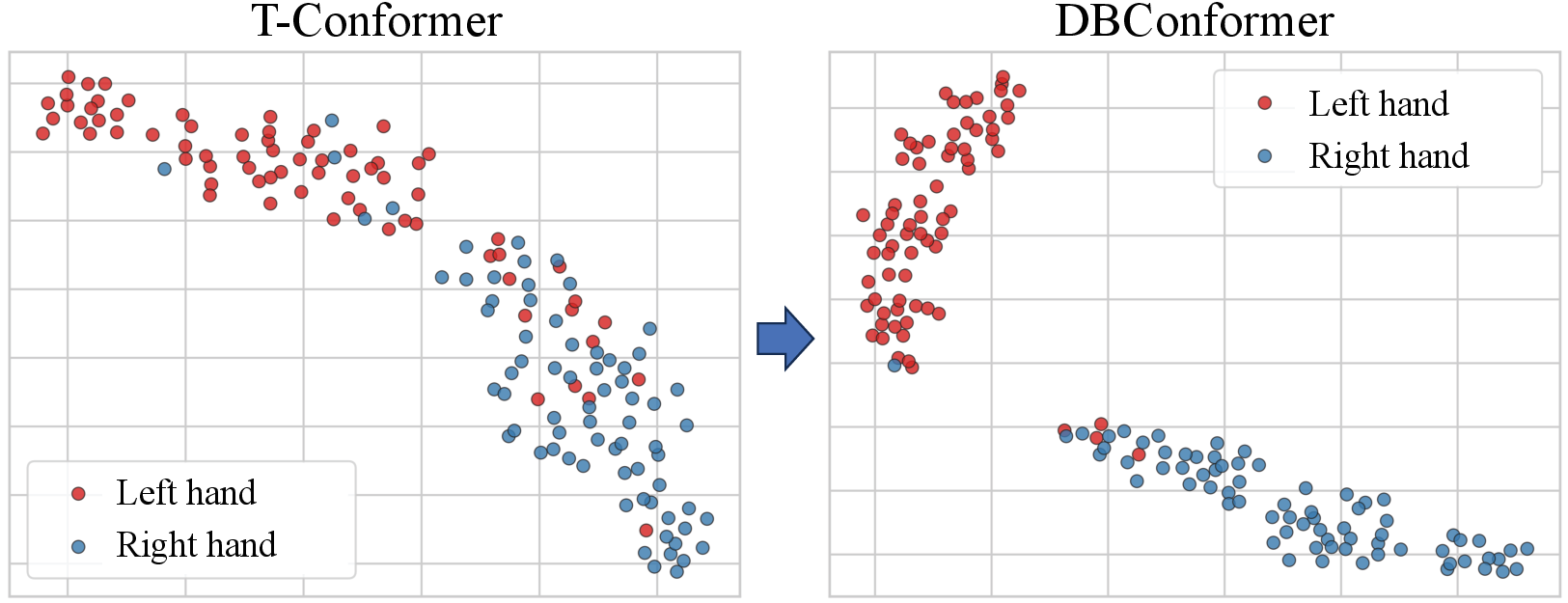}}
\subfigure[]{\includegraphics[width=.48\linewidth,clip]{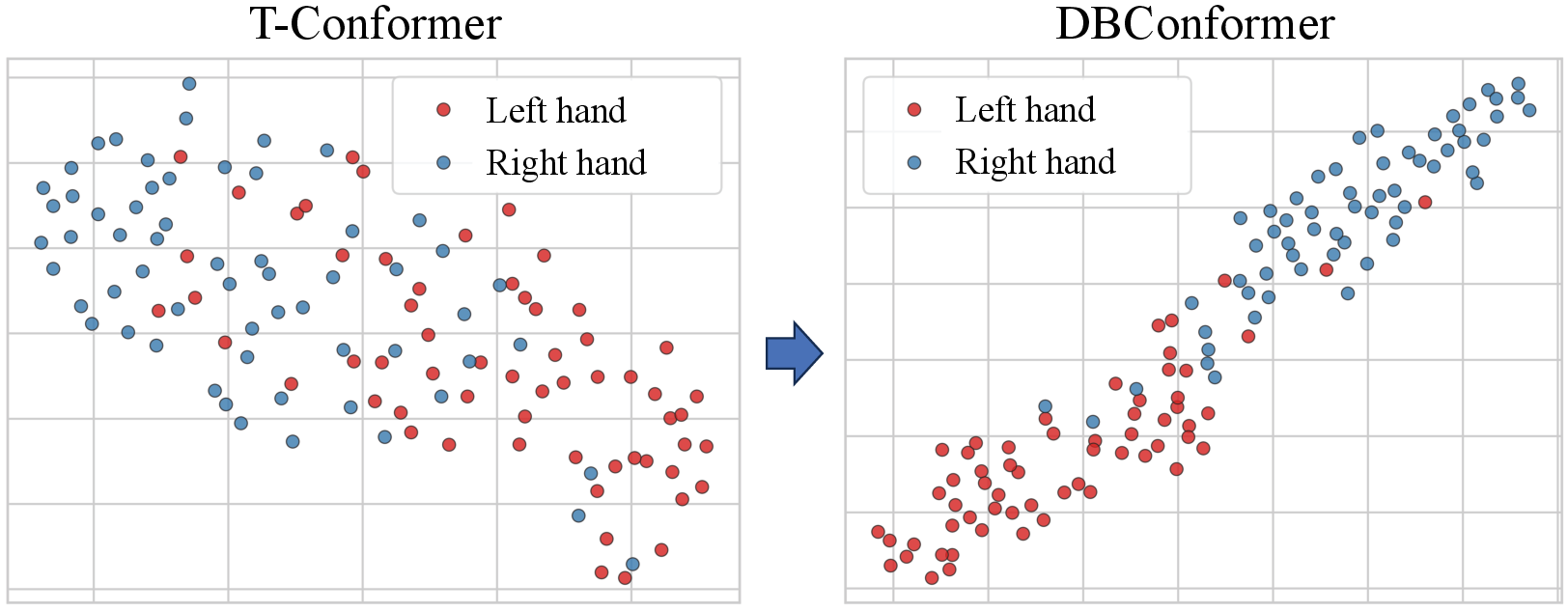}}
\subfigure[]{\includegraphics[width=.48\linewidth,clip]{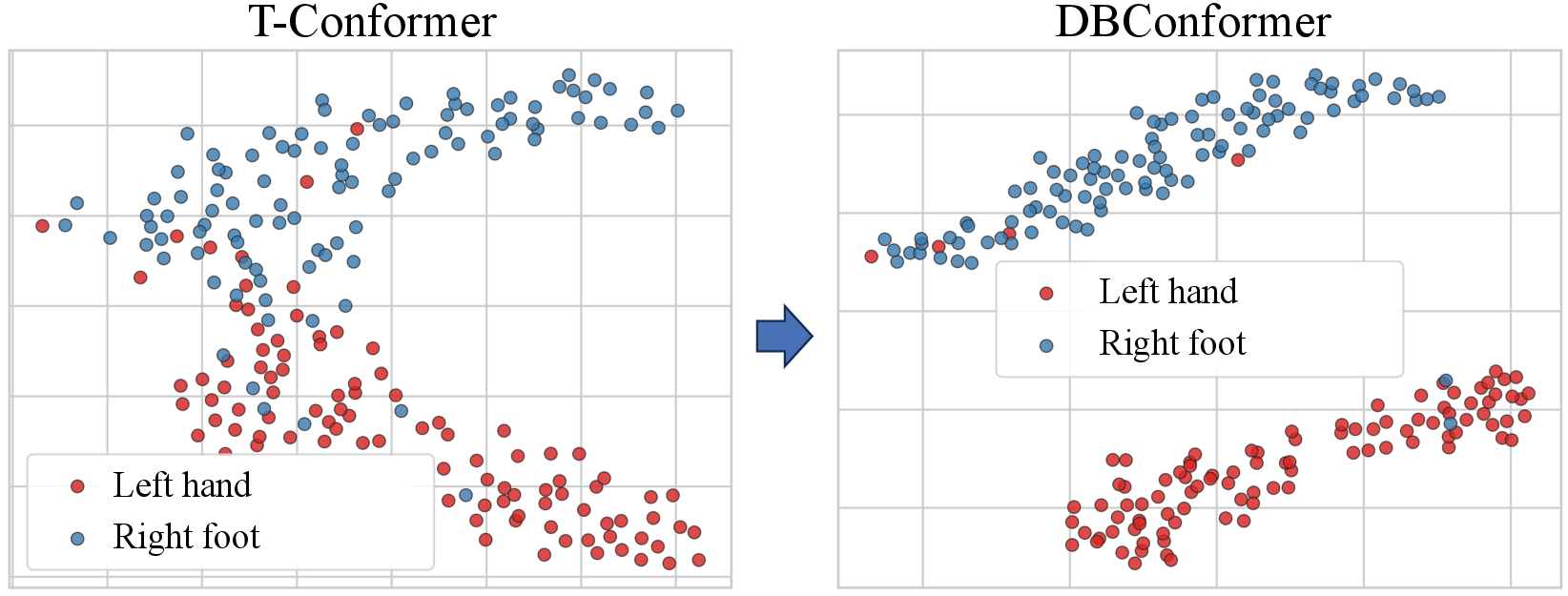}}
\subfigure[]{\includegraphics[width=.48\linewidth,clip]{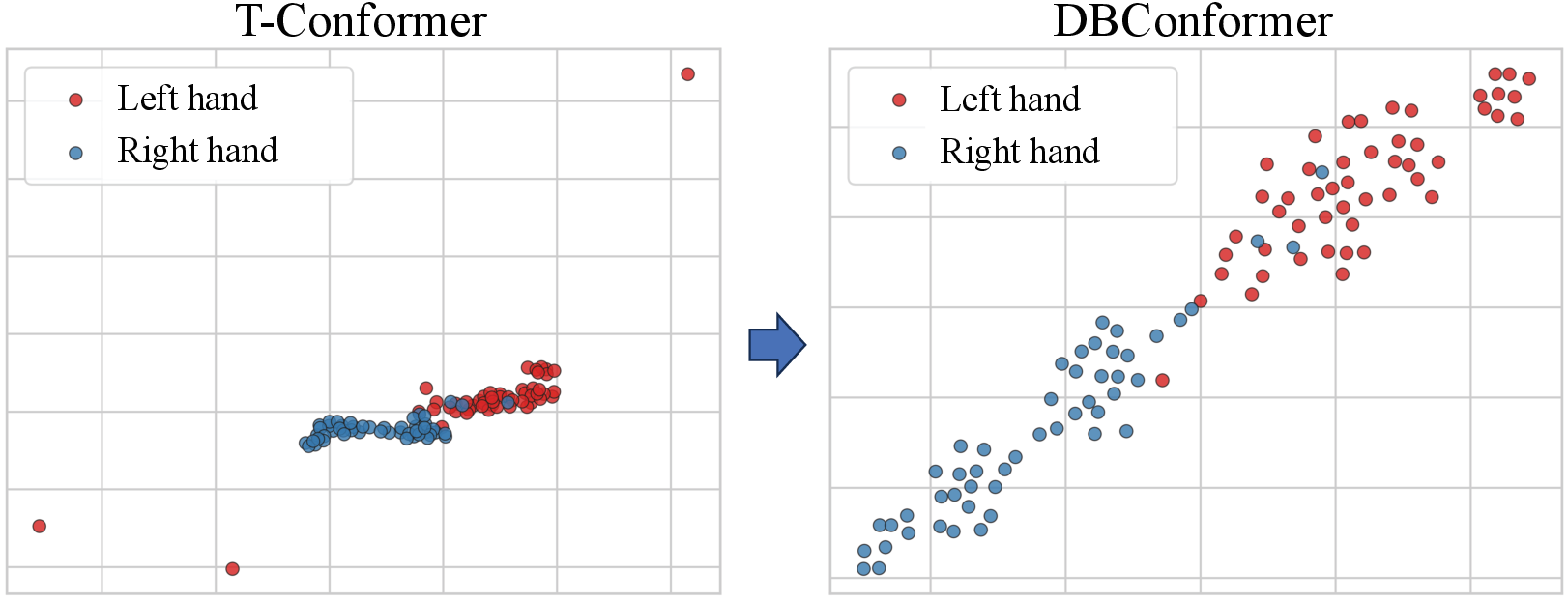}}
\caption{$t$-SNE visualizations of features extracted from T-Conformer and DBConformer on (a) BNCI2014001, (b) BNCI2014004, (c) Blankertz2007, and (d) Zhou2016 datasets. Different categories are encoded by colors.} \label{fig:tsne}
\end{figure*}

To further evaluate the impact of dual-branch architecture, we conducted feature visualization experiments using $t$-SNE \cite{VanderMaaten2008a}. Features extracted by T-Conformer (temporal branch only) and DBConformer (dual-branch) were compared on four MI datasets, shown in Fig.~\ref{fig:tsne}. Observe that:
\begin{itemize}
\item \textit{T-Conformer alone exhibited limited separability}. Although T-Conformer captured meaningful temporal structures, its extracted features tended to form overlapping clusters across classes in datasets such as BNCI2014004 and Blankertz2007, which limited class discriminability.
\item \textit{DBConformer extracted more structured and well-separated features}. With the integration of spatial encoding from the S-Conformer, DBConformer enhanced the inter-class margins and promotes cluster compactness across all datasets.
\item \textit{The improvement was consistent across various types of MI classification tasks}. The benefit of dual-branch spatio-temporal modeling was observable regardless of categories (e.g., hand vs. hand/foot), demonstrating the generalization and robustness of the proposed design.
\end{itemize}

\subsection{Visualization of Spatio-Temporal Self-Attention}
\begin{figure*}[htpb]\centering
\subfigure[]{\includegraphics[width=.9\linewidth,clip]{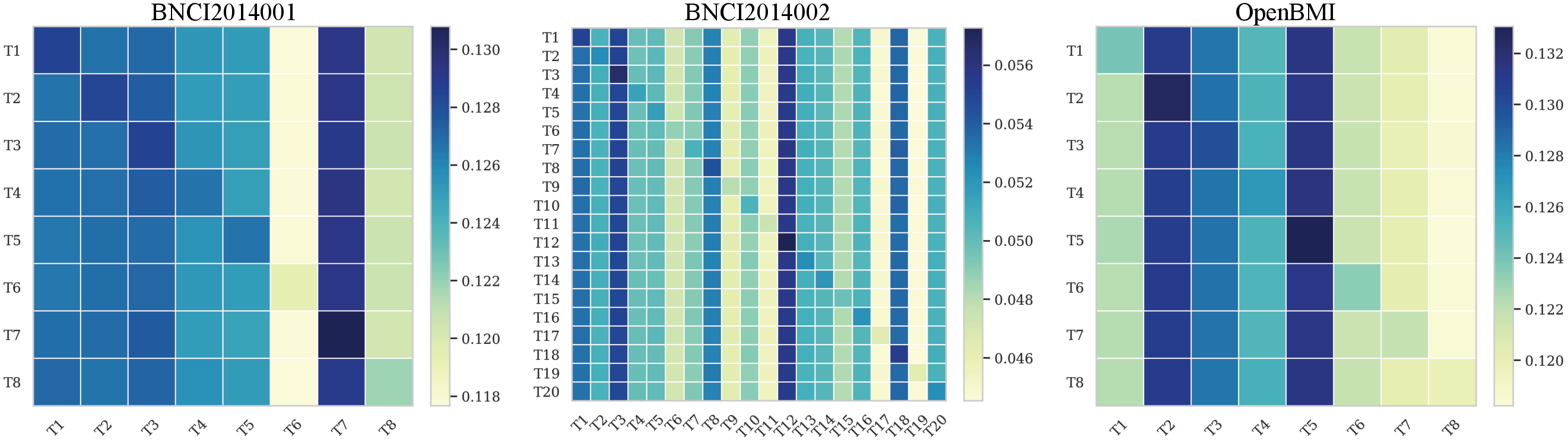}}
\subfigure[]{\includegraphics[width=.9\linewidth,clip]{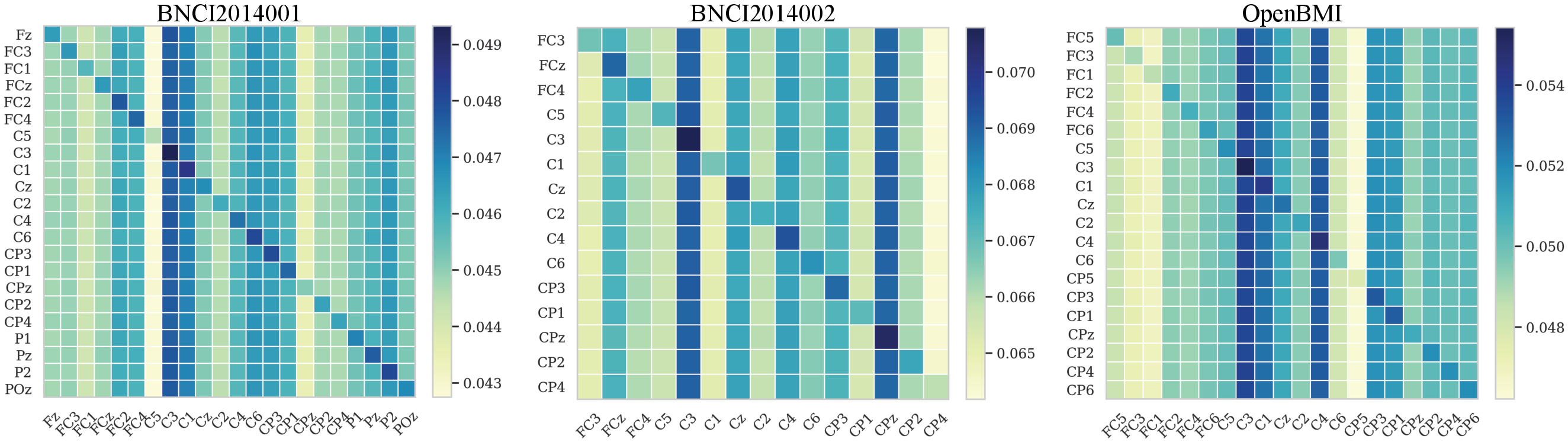}}
\subfigure[]{\includegraphics[width=.9\linewidth,clip]{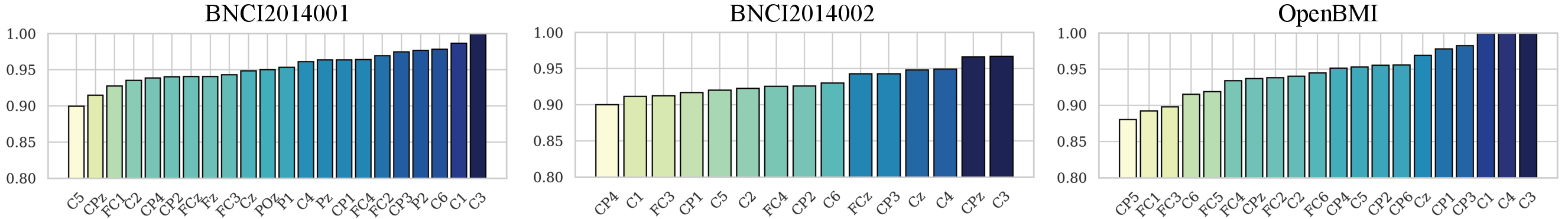}}
\caption{Visualizations of temporal and spatial self-attention across three MI datasets. (a) Temporal attention heatmap of time windows; (b) Spatial attention heatmap of EEG channels; (c) Channel importance of electrodes.} \label{fig:attn_vis}
\end{figure*}

To further examine the interpretability of DBConformer, Fig.~\ref{fig:attn_vis} visualized the self-attention matrices learned in both temporal and spatial branches on BNCI2014001, BNCI2014002, and OpenBMI datasets. Observe that:
\begin{itemize}
\item The temporal self-attention maps in Fig.~\ref{fig:attn_vis}(a) exhibited diagonal activations, which corresponded to the model’s focus on local neighboring time windows, as well as noticeable off-diagonal activations that captured long-range temporal dependencies. Across three datasets, DBConformer emphasized short-term EEG dynamics and integrated global temporal information, enabling a more comprehensive representation of MI.
\item The spatial self-attention maps in Fig.~\ref{fig:attn_vis}(b) revealed both diagonal activations corresponding to self-focus on individual channels, and distinct off-diagonal activations reflecting inter-channel interactions. In three datasets, channels over the sensorimotor cortex (e.g., C3, Cz, C4) exhibited the top attention intensities, indicating that DBConformer effectively captured physiologically meaningful intra-channel and inter-channel dependencies.
\item Furthermore, channel importance based on incoming off-diagonal attention weights is quantified in Fig.~\ref{fig:attn_vis}(c). Motor-related channels consistently dominated the importance ranking, confirming that DBConformer robustly identifies physiologically relevant channels.
\end{itemize}

\subsection{Model Complexity and Performance Analysis}
Table~\ref{tab:model_complexity} presents a comparative analysis of trainable parameter counts, training duration, and classification accuracy across DBConformer and nine baseline models. Observe that:
\begin{itemize}
\item Compared to lightweight CNN models, DBConformer introduced additional complexity due to the dual branches. Nonetheless, the trade-off in model size and training time was well-compensated by superior performance.
\item DBConformer achieved the highest classification accuracy, surpassing all competing baseline models. Despite incorporating a dual-branch Conformer design, it maintained a moderate number of parameters, with over 8$\times$ fewer than the high-capacity EEG Conformer.
\item IFNet and EEGNet achieved the shortest inference times among CNN-based models. Importantly, DBConformer achieved the fastest inference speed among CNN-Transformer hybrids, requiring only 8.7 ms per batch, which meets the real-time requirements for online BCI applications.
\end{itemize}

\begin{table*}[h]
\centering
\setlength{\tabcolsep}{0.2mm}
\renewcommand\arraystretch{1}
\footnotesize
\caption{Comparison of model complexity and performance metrics across DBConformer and twelve baseline models on the BNCI2014001 dataset.}
\label{tab:model_complexity}
\begin{tabular}{c|cccccc|c|cccccc}
\toprule
\multirow{2.5}{*}{Metric} & \multicolumn{6}{c|}{CNN} & CNN-Mamba & \multicolumn{6}{c}{CNN-Transformer} \\
\cmidrule{2-14}
 & EEGNet & SCNN & DCNN & FBCNet & ADFCNN & IFNet & SlimSeiz & CTNet & EEG Confor. & MSCFor. & TMSA-Net & MSVTNet & DBConformer \\
\midrule
\# Model Parameters & \textbf{1,406} & 46,084 & 320,479 & 7,042 & \underline{4,322} & 7,748 & 27,650 & \underline{27,284} & 789,668 & 150,626 & \textbf{17,869} & 72,892 & 92,066\\
Training Duration (s) & 73.64 & 76.44 & 90.97 & \underline{70.45} & 97.84 & \textbf{61.89} & 88.89 & \underline{105.95} & 223.36 & 151.63 & \textbf{96.33} & 159.80 & 203.79 \\
Inference Latency (s) & \underline{4.5\,$e^{-3}$} & 1.3\,$e^{-2}$ & 4.8\,$e^{-3}$ & 1.3\,$e^{-2}$ & 7.8\,$e^{-3}$ & \textbf{4.2\,$e^{-3}$} & 1.7\,$e^{-2}$ & 1.3\,$e^{-2}$ & 1.9\,$e^{-2}$ & 1.6\,$e^{-2}$ & \underline{8.8\,$e^{-3}$} & 9.0\,$e^{-3}$ & \textbf{8.7\,$e^{-3}$} \\
Accuracy (\%) & \underline{73.64} & 72.22 & 73.21 & 72.56 & 71.76 & \textbf{74.52} & 69.62 & 73.40 & 73.07 & 75.00 & \underline{76.67} & 74.60 & \textbf{77.67} \\
\bottomrule
\end{tabular}
\end{table*}

\subsection{Sensitivity Analysis on Architectural Design}
We further conducted a sensitivity analysis to explore how architectural design affects the DBConformer performance.

\subsubsection{Effect of Depth of T-Conformer and S-Conformer}
We varied the number of Transformer encoder layers in both T-Conformer and S-Conformer from 1 to 8, and evaluated the classification accuracy on BNCI2014001 and Zhou2016 datasets. Results in Fig.~\ref{fig:depth_heads}(a) showed that performance does not monotonically increase with depth. Instead, both datasets achieve the best performance with two Transformer layers. When the depth exceeds 4 layers, accuracy tends to decrease slightly, suggesting that excessive stacking of attention layers may lead to overfitting on the limited training samples in MI datasets. In contrast, shallow configurations achieve a favorable trade-off between model complexity and generalization.

\subsubsection{Effect of the Number of Attention Heads}
\begin{figure*}[h] \centering
\subfigure[]{\includegraphics[width=.4\linewidth,clip]{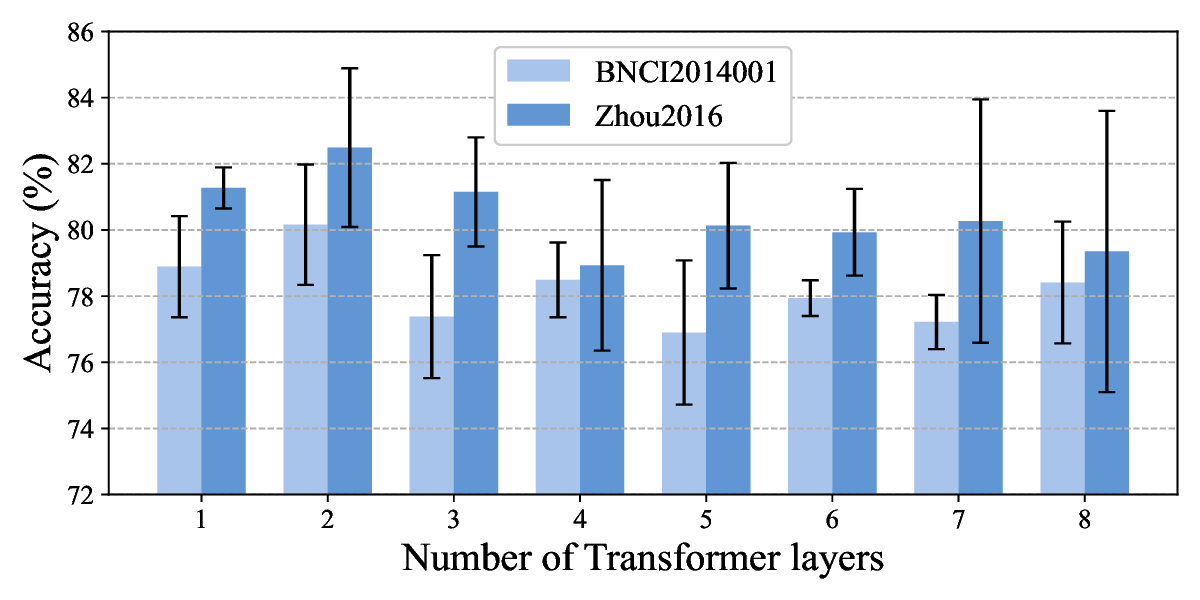}}
\subfigure[]{\includegraphics[width=.4\linewidth,clip]{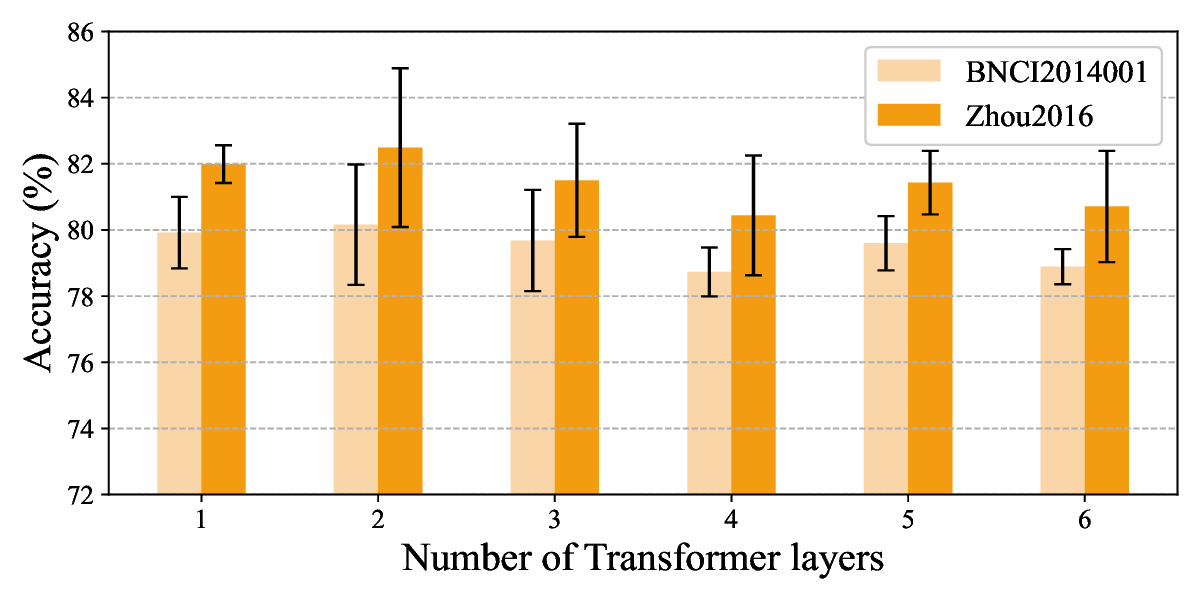}}
\caption{Sensitivity analysis on architectural design: (a) Number of Transformer layers in T-Conformer and S-Conformer, (b) Number of attention heads in T-Conformer and S-Conformer. Each bar represents the mean accuracy across subjects, with error bars denoting the standard deviation.} \label{fig:depth_heads}
\end{figure*}

Constrained by the embedding size of $D=40$, which requires the number of heads to be a divisor of $40$, we evaluated the number of attention heads ${1, 2, 4, 5, 8, 10}$ on BNCI2014001 and Zhou2016 datasets. As shown in Fig.~\ref{fig:depth_heads}(b), the results remain relatively stable across different settings, with DBConformer consistently outperforming baseline models. Notably, $2$ heads achieved the highest accuracy on both datasets. Increasing the number of heads beyond $4$ did not yield further improvements and sometimes led to a slight decrease in performance, suggesting that excessive partitioning of the embedding dimension may dilute feature representations.

\section{Conclusion}\label{sect:conclusions}
This paper proposed DBConformer, a dual-branch convolutional Transformer network tailored for EEG decoding. It integrates a T-Conformer to model long-range temporal dependencies and an S-Conformer to enhance inter-channel interactions, capturing both temporal dynamics and spatial patterns in EEG data.  Comprehensive experiments confirmed its effectiveness. In future work, DBConformer can be extended to support multi-view information fusion and plug-and-play real-time BCIs. Its superior performance and interpretability make it reliable for robust and explainable EEG decoding.

\bibliographystyle{IEEEtran} \bibliography{dbconformer}
\end{document}